\documentclass[10pt,twocolumn,letterpaper]{article}

\usepackage{cvpr}
\usepackage{times}
\usepackage{epsfig}
\usepackage{graphicx}
\usepackage{amsmath}
\usepackage{amssymb}
\usepackage{multirow}
\newcommand{\todo}[1]{}
\renewcommand{\todo}[1]{{\color{red} CS: {#1}}}

\usepackage{amsmath}

\usepackage{subcaption} % needed for figures
\usepackage{caption} % needed for figures

\usepackage[pagebackref=true,breaklinks=true,letterpaper=true,colorlinks,bookmarks=false]{hyperref}

\cvprfinalcopy % *** Uncomment this line for the final submission

\def\cvprPaperID{6433} % *** Enter the CVPR Paper ID here

\begin{document}

%%%%%%%%% TITLE
\title{ StereoDRNet: Dilated Residual Stereo Net}

%\maketitle

\twocolumn[{%
\renewcommand\twocolumn[1][]{#1}%
\maketitle
\vspace{-1.0cm}
\begin{center}
		{
		\vspace{-1.0cm}
		\normalsize
		Rohan Chabra $^{1 \dag}$~~~~~~Julian Straub $^{2}$~~~~~~Chris Sweeney$^2$~~~~~~Richard Newcombe$^2$ ~~~~~~Henry Fuchs$^1$\\ \vspace{0.1cm}
		
		$^1$University of North Carolina at Chapel Hill~~~~~~$^2$Facebook Reality Labs \vspace{0.05cm}
		
		\small{$^1$\{rohanc, fuchs\}@cs.unc.edu~~~~~~$^2$julian.straub@oculus.com,  \{sweeneychris, richard.newcombe\}@fb.com}
		}
    \centering
\includegraphics[width=\linewidth]{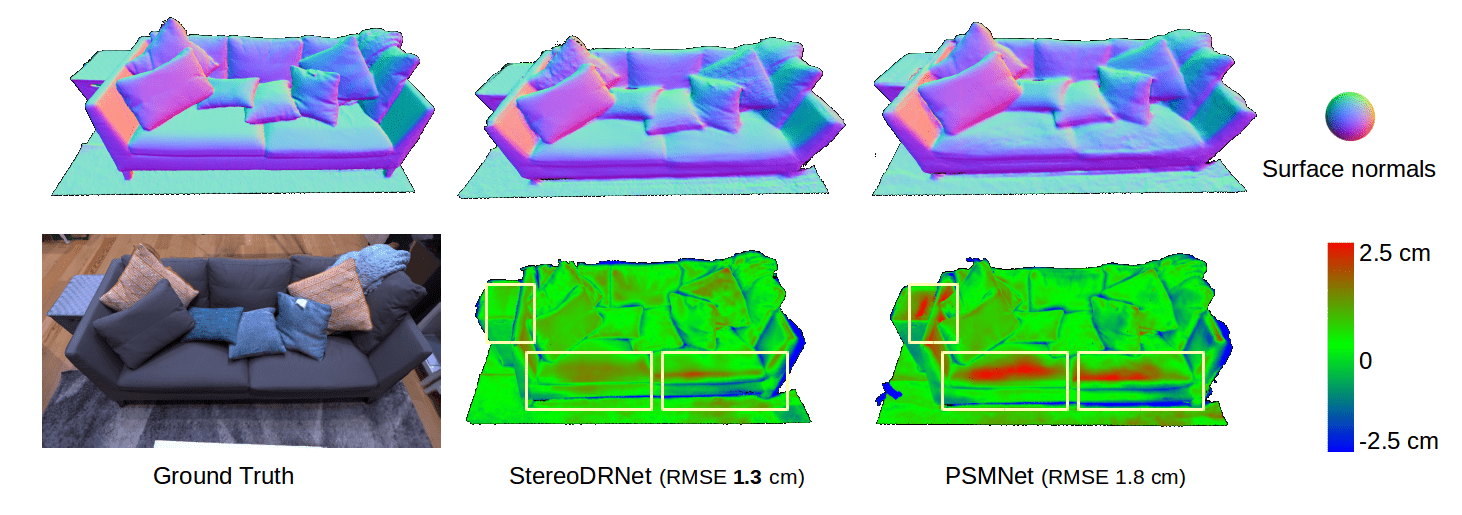}
    \captionof{figure}{
    StereoDRNet enables estimation of high quality depth maps that opens the door to high quality reconstruction by passive stereo video. In this figure we compare the output from dense reconstruction \cite{newcombe2011kinectfusion} built form depth maps generated by StereoDRNet, PSMNet~\cite{chang2018pyramid} and a structured light system \cite{whelan2018reconstructing} (termed Ground Truth). 
    We report and visualize point-to-plane distance RMS error on the reconstructed meshes with respect to the ground truth demonstrating the improvement in reconstruction over the state-of-the-art.}\label{fig:teaser}
\end{center}%
}]

%%%%%%%%% ABSTRACT
\begin{abstract}
\vspace{-0.35cm}
We propose a system that uses a convolution neural network (CNN) to estimate depth from a stereo pair followed by volumetric fusion of the predicted depth maps to produce a 3D reconstruction of a scene. Our proposed depth refinement architecture, predicts view-consistent disparity and occlusion maps that helps the fusion system to produce geometrically consistent reconstructions. We utilize 3D dilated convolutions in our proposed cost filtering network that yields better filtering while almost halving the computational cost in comparison to state of the art cost filtering architectures.
For feature extraction we use the Vortex Pooling architecture ~\cite{xie2018vortex}.
The proposed method achieves state of the art results in KITTI 2012, KITTI 2015 and ETH 3D stereo benchmarks.
Finally, we demonstrate that our system is able to produce high fidelity 3D scene reconstructions that outperforms the state of the art stereo system. 
   
\end{abstract}

\let\thefootnote\relax\footnotetext{\dag Work performed during internship at Facebook Reality Labs.
}
\vspace{-0.35cm}
\section{Introduction}
  Depth from stereo vision has been heavily studied in computer vision field for the last few decades. Depth estimation has various applications in autonomous driving, dense reconstruction and 3D objects and human tracking. Virtual Reality and Augmented Reality systems require depth estimations to build dense spatial maps of the environment for interaction and scene understanding. For proper rendering and interaction between virtual and real objects in an augmented 3D world, the depth is expected to be both dense and correct around object boundaries. Depth sensors such as structured light and time of flight sensors are often used to build such spatial maps of indoor environments. These sensors often use illumination sources which require power and space that exceeds the expected budget of an envisioned AR system. Since these sensors use infrared vision, they do not work well in bright sun light environment or in presence of other infrared sources. \\
%   They often interfere with each other which is a disadvantage for a collaborative ecosystem. Depth sensors in general have problems detecting reflective metal surfaces, Time of Flight sensors suffer from multi-path interference, and structured light sensors often have problems in detecting thin structures and sharp object boundaries and corners due to sparsity of their projected patterns for depth extraction.
  On the other hand, the depth from stereo vision systems have a strong advantage of working in both indoors and in sunlight environments. Since these systems use passive image data, they do not interfere with each other or with the environment materials. Moreover, the resolution of passive stereo systems is typically greater than the sparse patterns used in structured light depth sensors, so these methods have capabilities to produce depth with accurate object boundaries and corners. Due to recent advancements in camera and mobile technology the image sensors have dramatically reduced in size and have significantly improved in resolution and image quality. All these qualities makes passive stereo system a better fit for being a depth estimator for a AR or VR system. However, stereo systems have their own disadvantages, such as ambiguous predictions in texture-less or repeating/confusing textured surfaces. In order to deal with these homogeneous regions traditional methods make use of handcrafted functions and optimize the parameters globally on the entire image. Recent methods use machine learning to derive the functions and it's parameters from the data that is used in training. As these functions tend to be highly non-linear, they tend to yield reasonable approximations even on the homogeneous and reflective surfaces.

Our key contributions are as follows: \\
\textbullet\ \textbf{Novel Disparity Refinement Network}: The main motivation of our work is to predict geometrically consistent disparity maps for stereo input that can be directly used by TSDF-based fusion system like KinectFusion~\cite{newcombe2011kinectfusion} for simultaneous tracking and mapping. Surface normals are an important factor in fusion weight computation in KinectFusion-like systems, and we observed that state of the art stereo systems such as PSMNet produces disparity maps that are not geometrically consistent which negatively affect TSDF fusion. To address this issue, we propose a novel refinement network  which takes geometric error $E_g$, photometric error $E_p$ and unrefined disparity as input and produces refined disparity (via residual learning) and the occlusion map. \\
\textbullet\ \textbf{3D Dilated Convolutions in Cost Filtering}: State of the art stereo systems such as PSMNet\cite{chang2018pyramid} and GC-Net\cite{kendall2017end} that use 3D cost filtering approach use most of the computational resources in the filtering module of their system. We observe that using 3D dilated convolutions in all three dimensions i.e (width, height, and disparity channels) in a structure shown in Fig.~\ref{fig:DRNFUll} gave us better results with less compute (refer to Table.\ref{Table:SceneFLow}).\\
% Thus, our proposed cost filtering approach brings the overall system closer to real-time and hence promises frequent updates of spatial maps for various applications.\\
\textbullet\ \textbf{Other Contributions}: We observe that Vortex Pooling compared to spatial pyramid pooling (used in PSMNet) provides better results (refer to ablation study \ref{Table:AbalationNetwork}). We found the exclusion masks used to filter non-confident regions of ground truth for fine-tuning our model as discussed in Sec ~\ref{sec:SceneReconstruction} to be very useful in obtaining sharp edges and fine details in disparity predictions.
 We achieve 1.3 - 2.1 cm RMSE on 3D reconstructions of three scenes that we prepared using structured light system proposed in \cite{whelan2018reconstructing}.\\

\section{Related Work}
  Depth from stereo has been widely explored in the literature, we refer interested readers to surveys and methods described in \cite{scharstein2002taxonomy}. Broadly speaking stereo matching can be categorized into computation of cost metrics, cost aggregation, global or semi-global optimization \cite{hirschmuller2008stereo} and refinement or filtering processes. Traditionally global cost filtering approaches used discrete labeling methods such as Graph Cuts \cite{kolmogorov2001computing} or used belief propagation techniques described in \cite{klaus2006segment} and \cite{bleyer2011patchmatch}. Total Variation denoising \cite{rudin1992nonlinear} has been used in cost filtering by methods described in \cite{zach2007globally}, \cite{newcombe2011dtam} and \cite{newcombe2012dense}.\\
  The state of the art in disparity estimation techniques use CNNs. MC-CNN~\cite{zbontar2016stereo} introduced a Siamese network to compare two image patches. The scores on matching was used along with the semi-global matching process \cite{hirschmuller2008stereo} to predict consistent disparity estimation. DispNet~\cite{mayer2016large} demonstrates an end-to-end disparity estimation neural network with a correlation layer (dot product of features) for stereo volume construction. Liang~et~al.~\cite{liang2018learning} improved DispNet by introducing novel iterative filtering process. GC-Net~\cite{kendall2017end} introduces a method to filter 4D cost using a 3D cost filtering approach and the soft argmax process to regress depth. PSMNet~\cite{chang2018pyramid} improved GC-Net by enriching features with better global context using pyramid spatial pooling process. They also show effective use of stacked residual networks in cost filtering process.\\
  Xie~et~al.~\cite{xie2018vortex} introduce vortex pooling which is an improvement of the atrous spatial pooling approach used in Deep lab \cite{chen2018deeplab}. Atrous pooling uses convolutions with various dilation steps to increase receptive fields of a CNN filter. The vortex pooling technique uses average pooling in grids of varying dimensions before dilated convolutions to utilize information from the pixels which were not used in bigger dilation steps. The size of average pool grids grows with the increase in dilation size. We use the feature extraction described in Vortex pooling and improve the cost filtering approach described by PSMNet.\\
  Our proposed refinement network takes geometric error $E_g$, photometric error $E_p$ and unrefined disparity as input and produces refined disparity (via residual learning) and the occlusion map. Refinement procedures proposed in CRL~\cite{pang2017cascade}, iResNet~\cite{liang2018learning}, StereoNet~\cite{khamis2018stereonet} and FlowNet2~\cite{ilg2017flownet} only use photometeric error (either in image or feature domain) as part of the input in the refinement networks. To the best of our knowledge we are the first to explore the importance of geometric error and occlusion training for disparity refinement.\\
 
\section{Algorithm}
  In this section we describe our architecture that predicts disparity for the input stereo pair. Instead of using a generic encoder-decoder CNN we break our algorithm into feature extraction, cost volume filtering and refinement procedures. \\
  \subsection{Feature Extraction}
\begin{figure}
  \centering
  		\includegraphics[width=0.8\linewidth]{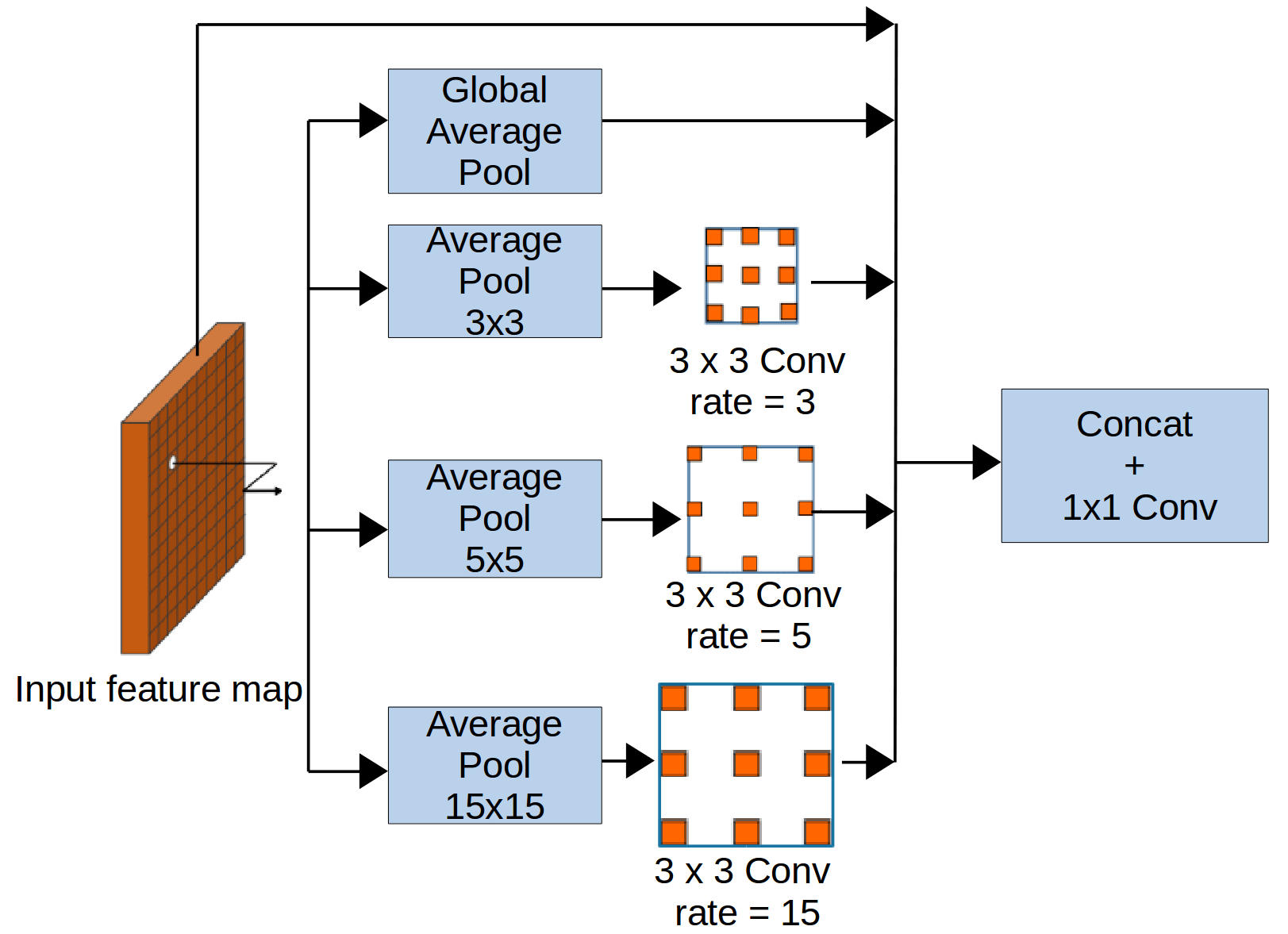}
  	\caption{StereoDRNet Vortex Pooling architecture derived from ~\cite{xie2018vortex}.} 
  	\label{fig:VortexPooling}
  \end{figure}

  The feature extraction starts with a small shared weight Siamese network which takes input as images and encodes the input to a set of features. As these features will be used for stereo matching we want them to have both local and global contextual information. To encode local spatial information in our feature maps we start by downsampling the input by use of convolutions with stride of 2. Instead of having a large $5\times 5$ convolution we use three $3\times 3$ filters where first convolution has stride of 2. We bring the resolution to a fourth by having two of such blocks. In order to encode more contextual information we choose Vortex Pooling \cite{xie2018vortex} on the learned local feature maps Fig.~\ref{fig:VortexPooling}. Each of our convolutions are followed by batch normalization and ReLU activation except on the last 3x3 convolution on the spatial pooling output. In order to keep the feature information compact we keep the feature dimension size as 32 throughout the feature extraction process.
  
%   \begin{figure*}
%   	\begin{center}
%   		\includegraphics[width=\textwidth]{figures/DRN_High}
%   	\end{center}
  	
%   	\caption{ This figures demonstrate the overall network architecture. We also demonstrate the Spatial Pooling architecture and two cost filtering approaches. } 
%   	\label{fig:DRNetStruct}
%   \end{figure*}
   \begin{figure*}
  	\centering
  		\includegraphics[width=1.0\textwidth,clip,trim=0 3 0 0]{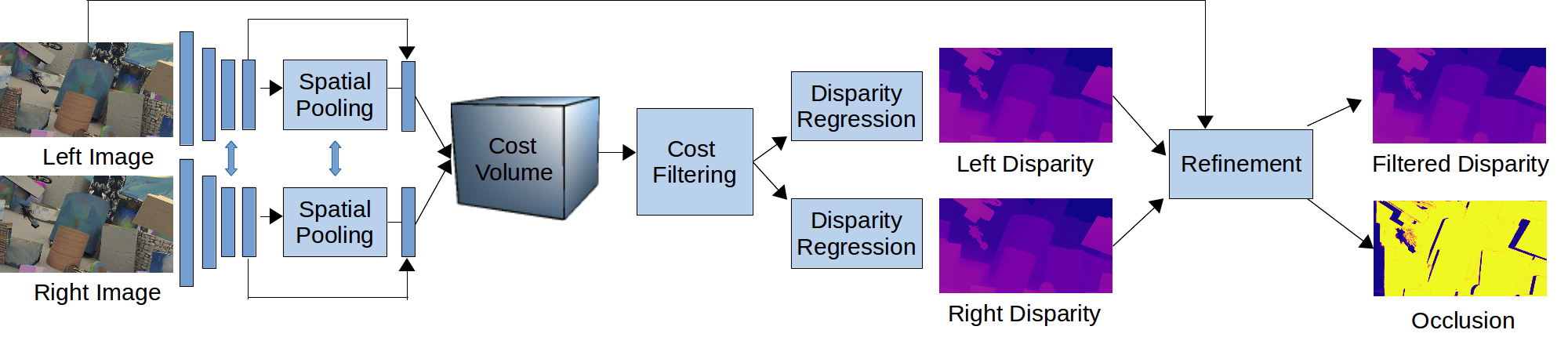}
  	\caption{StereoDRNet network architecture pipeline. } 
  	\label{fig:DRNPipeline}
  \end{figure*}
  
  \begin{figure*}
  \centering
  		\includegraphics[width=0.9\linewidth]{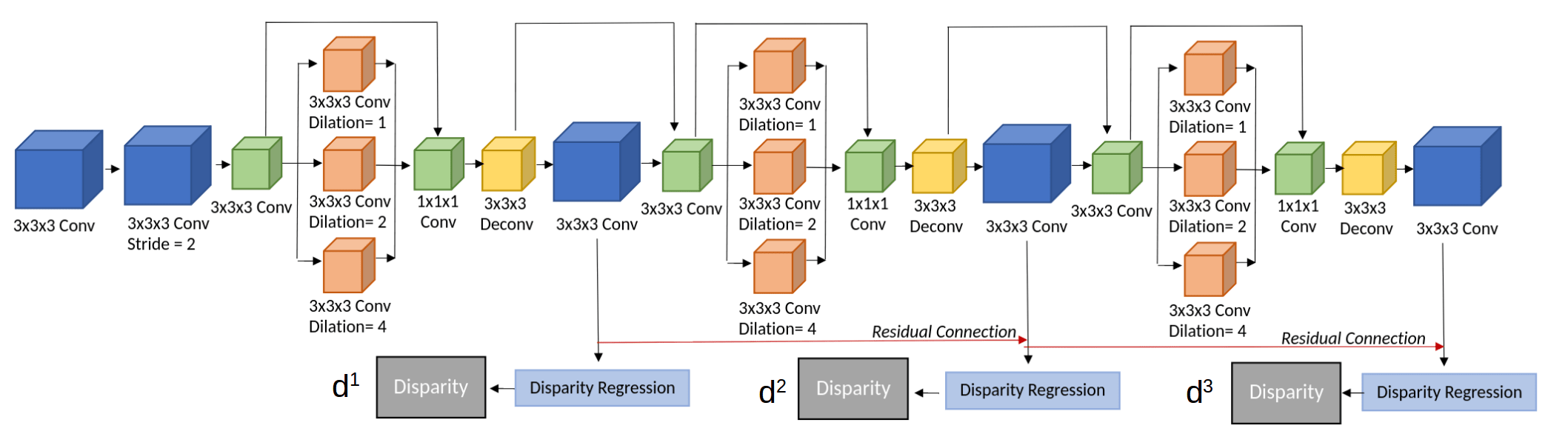}
  	\caption{Proposed dilated cost filtering approach with residual connections. } 
  	\label{fig:DRNFUll}
  \end{figure*}
  
 \subsection{Cost Volume Filtering}
 \label{sec:CostVol}
  We use the features extracted in the previous step to produce a stereo cost volume. While several approaches in the literature (\cite{kendall2017end},\cite{mayer2016large}) use concatenation or dot products of the stereo features to obtain the cost volume, we found simple arithmetic difference to be just as effective. 
  
  While the simple argmin on the cost should in principle lead to the correct local minimum solution, it has been shown several times in literature ~\cite{newcombe2011dtam}, \cite{hirschmuller2008stereo},\cite{scharstein2002taxonomy} that it is common for the solution to have several local minima. Surfaces with homogeneous or repeating texture are particularly prone to this problem. By posing the cost filtering as a deep learning process with multiple convolutions and non-linear activations we attempt to resolve these ambiguities and find the correct local minimum. 
  
  We start by processing our cost volume with a $3 \times 3 \times 3$ convolution along the width, height and depth dimensions. We then reduce the resolution of the cost by a convolution with stride of 2 followed by convolutions with dilation 1, 2, 4 in parallel. A convolution on the concatenation of the dilated convolution filters is used to combine the information fetched from varying receptive fields.
  
  Residual learning has been shown to be very effective in disparity refinement process so we propose a cascade of such blocks to iteratively improve the quality of our disparity prediction. We depict the entire cost filtering process as Dilated Residual Cost Filtering in Fig.~\ref{fig:DRNFUll}. In this figure notice how our network is designed to produce $k=3$ disparity maps labeled as $d^k$. 
  
  Our network architecture that supports refinement predicts disparities for both left and right view as separate channels in disparity predictions $d^{k}$. Note that we construct the cost for both left and right views and concatenate them before filtering; this ensures that the cost filtering method is provided with cost information for both views. Please refer to Table.~\ref{Table:Network} for exact architecture details.\\
  
%   \begin{figure}
%   	\begin{center}
%   		\includegraphics[width=1.0\linewidth]{figures/DRNCost}
%   	\end{center}
  	
%   	\caption{ This figures demonstrate the entire cost volume filtering procedure of our network where $d_1$, $d_2$ and $d_3$ are the disparity predictions.} 
%   	\label{fig:DRNCost}
%   \end{figure}
\vspace{-0.35cm}
  \subsection{Disparity Regression \label{sec:disparityRegression}}
  
  In order to have a differentiable argmax we use soft argmax as proposed by GC-Net \cite{kendall2017end}. For each pixel $i$ the regressed disparity estimation $d_i$ is defined as a weighted softmax function:
  \begin{equation}
%   d_i =  \tfrac{ d \,  e^{-C_i(d)}}{ \sum\limits_{d'=1}^{N}{d' e^{-C_i(d')}}} \,,
   d_i =  \sum\limits_{d=1}^{N}{d \, \tfrac{e^{-C_i(d)}}{ \sum\limits_{d'=1}^{N}{e^{-C_i(d')}}}} \,,
  \label{eqn:1}
  \end{equation}
  where $C_i$ is the cost at pixel $i$ and $N$ is the maximum disparity.
  The loss $L^k$ for each of the proposed disparity maps $d^k$ (as shown in Fig.~\ref{fig:DRNFUll}) in our dilated residual cost filtering architecture, relies on the Huber loss $\rho$ and is defined as:
  \begin{equation}
  L^k = \sum\limits_{i}^{M} \rho(d^{k}_i,\hat{d_i})\,,
  \label{eqn:2}
  \end{equation}
  where $d^{k}_i$ and $\hat{d_i}$ are the estimated and ground truth disparity at pixel $i$, respectively and $M$ is the total number of pixels. The total data loss $L_d$ is defined as:
  \begin{equation}
  L_d = \sum\limits_{k=1}^{3} w^{k}  L^{k}\,,
  \label{eqn:4}
  \end{equation}
  where $w^{k}$ is the weight for each disparity map $d^k$.

  \subsection{Disparity Refinement}
  In order to make the disparity estimation robust to occlusions and view consistency we further optimize the estimate. 
  For brevity we label the third disparity prediction $d^3$ ($k$ = 3) described in Sec.~\ref{sec:CostVol} for left view as $D_l$ and for right view as $D_r$. In our refinement network we warp the right image $I_r$ to left view via the warp $W$ and evaluate the image reconstruction error map $E_p$ for the left image $I_l$ as:
  \begin{equation}
  E_p = |I_l - W(I_r,D_r)| \,.
  \label{eqn:5}
  \end{equation}
  By warping $D_r$ to the left view and using the left disparity $D_l$ we can evaluate the geometric consistency error map $E_g$ as:  
  \begin{equation}
  E_g = |D_l - W(D_r,D_l)| \,.
  \label{eqn:6}
  \end{equation}
  
  \begin{figure}
  	\begin{center}
  		\includegraphics[width=0.95\linewidth]{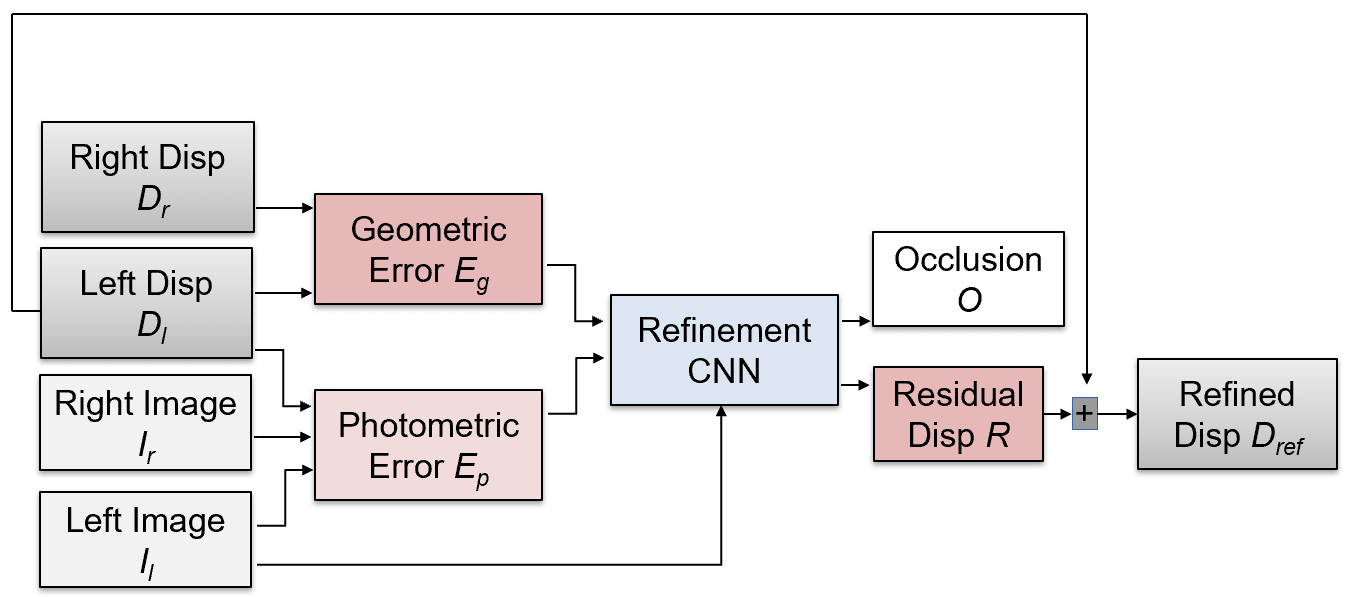}
  	\end{center}
  	
  	\caption{ StereoDRNet refinement architecture.} 
  	\label{fig:Refinement}
  \end{figure}

\begin{figure}[ht]
	\centering
		\includegraphics[width=1.0\linewidth]{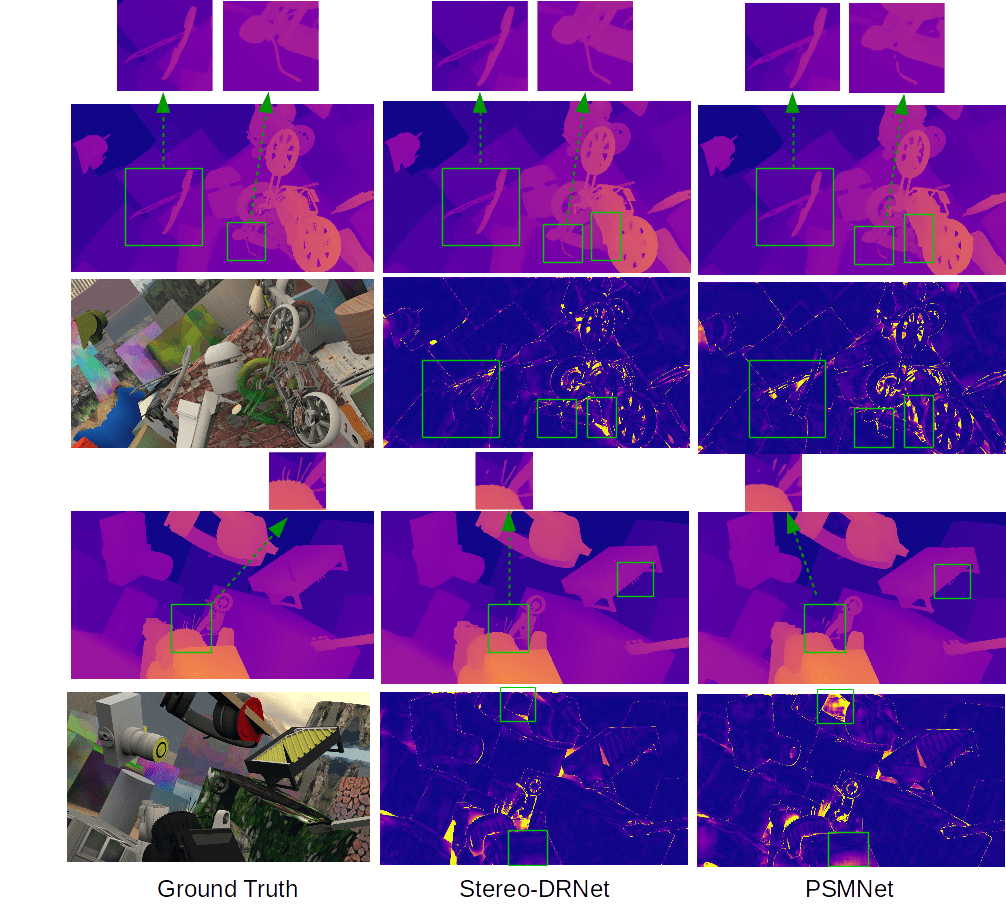}
	\caption{ Disparity prediction comparison between our network (Stereo-DRNet) and PSMNet~\cite{chang2018pyramid} on the SceneFlow dataset. The top row shows disparity and the bottom row shows the EPE map. Note how our network is able to recover thin and small structures and at the same times shows lower error in homogeneous regions.} 
	\label{fig:SceneFLowDRN}
\end{figure}
  
  While we could just reduce these error terms directly into a loss function, we observed significant improvement by using photo-metric and geometric consistency error maps as input to the refinement network as these error terms are only meaningful for non occluding pixels (only pixels for which the consistency errors can be reduced).
  
  Our refinement network takes as input left image $I_l$, left disparity map $D_l$, image reconstruction error map $E_p$ and geometric error map $E_g$. We first filter left image and reconstruction error and left disparity and geometric error map $E_g$ independently by using one layer of convolution followed by batch normalization. Both these results are then concatenated and followed by atrous convolution \cite{papandreou2015modeling} to sample from a larger context without increasing the network size. We used dilations with rate 1, 2, 4, 8, 1, and 1 respectively. Finally a single $3\times 3$ convolution without ReLU or batch normalization is used to output an occlusion map $O$ and a disparity residual map $R$. Our final refined disparity map is labeled as $D_{ref}$. We demonstrate our refinement network in Fig.~\ref{fig:Refinement} and provide exact architecture details in Table.~\ref{Table:RefinementNetwork}. 
  
  We compute the cross entropy loss on the occlusion map $O$ as $L_o$
  \begin{equation}
  L_o = H(O,\hat{O}) \,,
  \label{eqn:7}
  \end{equation}
  where $\hat{O}$ is the ground truth occlusion map.
 
  The refinement loss $L_r$ is defined as 
  \begin{equation}
  L_r = \sum\limits_{i}^{M} \rho(d^{r}_i,\hat{d_i}) \,,
  \label{eqn:8}
  \end{equation}
  where $d^{r}_i$ is the value for a pixel $i$ in our refined disparity map $D_{ref}$ and $M$ is the total number of pixels.
  
  Our total loss function $L$ is defined as 
  \begin{equation}
  L = L_d + \lambda_1  L_r + \lambda_2 L_o \,,
  \label{eqn:9}
  \end{equation}
  where $\lambda_1$ and $\lambda_2$ are scalar weights.

\subsection{Training}
We implemented our neural network code in PyTorch. We tried to keep the training of our neural network similar to one described in PSMNet~\cite{chang2018pyramid} for ease of comparison. We used Adam optimizer~\cite{kingma2014adam} with $\beta_1$ = 0.9 and $\beta_2$ = 0.999 and normalized the image data before passing it to the network. In order to optimize the training procedure we cropped the images to 512x256 resolution. For training we used a mini-batch size of 8 on 2 Nvidia Titan-Xp GPUs. We used $w^1$ = 0.2, $w^2$ = 0.4, $w^3$ = 0.6, $\lambda_1$ = 1.2 and $\lambda_2$ = 0.3 weights in our proposed loss functions Eq.~\ref{eqn:4} and Eq.~\ref{eqn:9}.

\section{Experiments}
   
We tested our architecture on rectified stereo datasets such as SceneFlow, KITTI 2012, KITTI 2015 and ETH3D. We also demonstrate the utility of our system in building 3D reconstruction of indoor scenes. See the supplementary section for additional visual comparisons.
   
%   \begin{figure}
%   	\begin{center}
%   		\includegraphics[width=1.0\linewidth]{figures/Consistency}
%   	\end{center}
   	
%   	\caption{ This Fig.~demonstrate the results with and without refinement. We show the 3D mesh generated from individual depth maps. Note the refinement introduces better shapes because we condition our refinement network on geometric and photo-metric consistencies. } 
%   	\label{fig:Consistency}
%   \end{figure}
   
\subsection{SceneFlow Dataset}
   SceneFlow~\cite{mayer2016large} is a synthetic dataset with over $30,000$ stereo pairs for training and around $4000$ stereo pairs for evaluation. We use both left and right ground truth disparities for training our network. We compute the ground truth occlusion map by defining as occluded any pixel with disparities inconsistency larger than 1 px. This dataset is challenging due to presence of occlusions, thin structures and large disparities.  
   
   In Fig.~\ref{fig:SceneFLowDRN} we visually compare our results with PSMNet~\cite{chang2018pyramid}. Our system infers better structural details in the disparity image and also produces consistent depth maps with significantly less errors in homogeneous regions. We further visualize the effect of our refinement network in Fig.~\ref{fig:NormalSceneFlow}.
%   Fig.~\ref{fig:Consistency}. 
%   The 3D mesh generated from depth maps predicted using architectures with refinement network as described in Sec.~\ref{sec:disparityRegression} is visually cleaner. Since we condition our refinement network on geometric and photo-metric consistencies our disparity predictions recover better objects shapes.

Table~\ref{Table:SceneFLow} shows a quantitative analysis of our architecture with and without refinement network. Stereo-DRNet achieves significantly lower end point error while reducing computation time. Our proposed cost filtering approach achieves better accuracy with significantly less compute, demonstrating the effectiveness of the proposed dilated residual cost filtering approach.
   
\noindent\textbf{Ablation study:} In Table~\ref{Table:AbalationNetwork} we show a complete EPE breakdown for different parts of our network on the SceneFlow dataset. Both vortex pooling and refinement procedure add marginal performance gains. Co-training occlusion map with residual disparity drastically improves the mean end point disparity error of the final disparity from 0.93 px to 0.86 px. Passing only the photometric error into the refinement network actually degrades the performance.
   
   \begin{table}[h]
   	\centering
   	\footnotesize \setlength\tabcolsep{2.9pt} \renewcommand{\arraystretch}{1.2}
   		\begin{tabular}{|c|c|c|c|c|}   			\hline
   			Method & EPE & Total FLOPS & 3D-Conv FLOPS & FPS \\
   	% 		\hline DispnetC\cite{mayer2016large} & 1.68 & - & - & \textbf{15.0} \\
   			\hline CRL\cite{pang2017cascade} & 1.32 & - & - & 2.1 \\
   			\hline GC-Net\cite{kendall2017end} & 2.51 & 8789 GMac & 8749 GMac & 1.1 \\
   			\hline PSMNet\cite{chang2018pyramid} & 1.09 & 2594 GMac & 2362 GMac & 2.3 \\
   			\hline Ours & 0.98 & \textbf{1410 GMac} & \textbf{1119 GMac} & \textbf{4.3} \\
   			\hline Ours-Ref & \textbf{0.86} & 1711 GMacs & 1356 GMacs & 3.6 \\
   		
   			\hline
   		\end{tabular}
   	\caption{Quantitative comparison of the proposed Stereo-DRNet with the state of the art methods on the SceneFlow dataset. EPE represent the mean end point error in disparity. FPS and FLOPS (needed by the convolution layers) are measured on full $960\times 540$ resolution stereo pairs. Notice even our unrefined disparity architecture outperforms the state of the art method PSMNet \cite{chang2018pyramid} while requiring significantly less computation.}
   	\label{Table:SceneFLow}
   \end{table}
   
  \begin{table}[h]
    	\centering
    	\footnotesize \setlength\tabcolsep{3.2pt} \renewcommand{\arraystretch}{1.2}
    		\begin{tabular}{|c|c|c|c|c|c|c|c|c|}
    			\hline\multicolumn{7}{|c|}{Network Architecture} &  \multicolumn{1}{|c|}{SceneFlow}  & \multicolumn{1}{|c|}{KITTI-2015} \\
    			\hline\multicolumn{1}{|c|}{\multirow{2}{*}{Pooling}} & \multicolumn{3}{|c|}{Cost Filtering} & \multicolumn{3}{|c|}{Refinement}  & \multicolumn{1}{|c|}{\multirow{2}{*}{EPE}} & \multicolumn{1}{|c|}{\multirow{2}{*}{Val Error(\%)}} \\ \cline{2-7}
    			
    			\multicolumn{1}{|c|}{} & \multicolumn{1}{|c|}{$d^1$} & \multicolumn{1}{|c|}{$d^2$} & \multicolumn{1}{|c|}{$d^3$} & \multicolumn{1}{|c|}{$E_p$} & \multicolumn{1}{|c|}{$E_g$} & \multicolumn{1}{|c|}{$L_o$} & \multicolumn{1}{|c|}{} & \multicolumn{1}{|c|}{}\\ \cline{1-9}
    			
		         Pyramid & \checkmark &  &  &  &  &  & 1.17 & 2.28 \\\cline{1-9}
		         Vortex & \checkmark &  &  &  &  &  & 1.13 & 2.14 \\\cline{1-9}
		         Vortex & \checkmark & \checkmark &  &  &  &  & 0.99 & 1.88 \\\cline{1-9}
		         Vortex & \checkmark & \checkmark & \checkmark &  &  &  & 0.98 & \textbf{1.74} \\\cline{1-9}
		         Pyramid & \checkmark & \checkmark & \checkmark &  &  &  & 1.00 & 1.81 \\\cline{1-9}
		         Vortex & \checkmark & \checkmark & \checkmark & \checkmark  &  &  & 1.03 & - \\\cline{1-9}
		         Vortex & \checkmark & \checkmark & \checkmark &   & \checkmark  & & 0.95 & - \\\cline{1-9}
		         Vortex & \checkmark & \checkmark & \checkmark & \checkmark  & \checkmark &  & 0.93 & - \\\cline{1-9}
		         Vortex & \checkmark & \checkmark & \checkmark & \checkmark  & \checkmark & \checkmark  & \textbf{0.86} & - \\\cline{1-9}
		         Pyramid & \checkmark & \checkmark & \checkmark & \checkmark  & \checkmark & \checkmark  & 0.96 & - \\\cline{1-9}

    		\end{tabular}
    	\caption{Ablation study of network architecture settings on SceneFlow and KITTI-2015 evaluation dataset.}
    	\label{Table:AbalationNetwork}
    \end{table}

     \begin{table}[h]
    	\centering
    	\footnotesize \setlength\tabcolsep{2.9pt} \renewcommand{\arraystretch}{1.2}
    		\begin{tabular}{|c|c|c|c|c|c|c|c|}
    			\hline\multicolumn{1}{|c|}{\multirow{2}{*}{Method}} & \multicolumn{2}{|c|}{\multirow{1}{*}{2px}} & \multicolumn{2}{|c|}{\multirow{1}{*}{3px}} &  \multicolumn{2}{|c|}{\multirow{1}{*}{Avg Error}}  & \multicolumn{1}{|c|}{\multirow{2}{*}{Time(s)}} \\\cline{2-7}
    			
    			\multicolumn{1}{|c|}{} & \multicolumn{1}{|c|}{Noc} & \multicolumn{1}{|c|}{All} & \multicolumn{1}{|c|}{Noc} & \multicolumn{1}{|c|}{All} & \multicolumn{1}{|c|}{Noc} & \multicolumn{1}{|c|}{All} & \multicolumn{1}{|c|}{} \\\cline{1-8}
    			
    			GC-NET\cite{kendall2017end} & 2.71 & 3.46 & 1.77 & 2.30 & 0.6 & 0.7 & 0.90  \\\cline{1-8}
    			EdgeStereo\cite{song2018edgestereo} & 2.79 & 3.43 & 1.73 & 2.18 & 0.5 & 0.6 & 0.48  \\\cline{1-8}
    			PDSNet\cite{tulyakov2018practical} & 3.82 & 4.65 & 1.92 & 2.53 & 0.9 & 1.0 & 0.50  \\\cline{1-8}
    			SegStereo\cite{yang2018segstereo} & 2.66 & 3.19 & 1.68 & 2.03 & 0.5 & 0.6 & 0.60  \\\cline{1-8}
    		    PSMNet\cite{chang2018pyramid} & 2.44 & 3.01 & 1.49 & 1.89 & 0.5 & 0.6 & 0.41 \\\cline{1-8}
    		    Ours & \textbf{2.29} & \textbf{2.87} & \textbf{1.42} & \textbf{1.83} & \textbf{0.5} & \textbf{0.5} & \textbf{0.23} \\\cline{1-8}
    		\end{tabular}
    	\caption{Comparison of disparity estimation from StereoDRNet with state of the art published methods on KITTI 2012 dataset.}
    	\label{Table:KITT2012}
    \end{table}
    
    \begin{table}[h]
    	\centering
    	 \footnotesize \setlength\tabcolsep{2.9pt} \renewcommand{\arraystretch}{1.2}
    		\begin{tabular}{|c|c|c|c|c|c|c|c|}
    			\hline\multicolumn{1}{|c|}{\multirow{2}{*}{Method}} & \multicolumn{3}{|c|}{\multirow{1}{*}{All(\%)}} & \multicolumn{3}{|c|}{\multirow{1}{*}{Noc(\%)}}   & \multicolumn{1}{|c|}{\multirow{2}{*}{Time(s)}} \\\cline{2-7}
    			
    			\multicolumn{1}{|c|}{} & \multicolumn{1}{|c|}{D1-bg} & \multicolumn{1}{|c|}{D1-fg} & \multicolumn{1}{|c|}{D1-all} & \multicolumn{1}{|c|}{D1-bg} & \multicolumn{1}{|c|}{D1-fg} & \multicolumn{1}{|c|}{D1-all} & \multicolumn{1}{|c|}{} \\\cline{1-8}
				DN-CSS\cite{ilg2018occlusions} & 2.39 & 5.71 & 2.94 & 2.23 & 4.96 & 2.68 &  \textbf{0.07}  \\\cline{1-8}
    			GC-NET\cite{kendall2017end} & 2.21 & 6.16 & 2.87 & 2.02 & 5.58 & 2.61 & 0.90  \\\cline{1-8}
    			CRL\cite{pang2017cascade} & 2.48 & \textbf{3.59} & 2.67 & 2.32 & \textbf{3.12} & 2.45 & 0.47  \\\cline{1-8}
    			EdgeStereo\cite{song2018edgestereo} & 2.27 & 4.18 & 2.59 & 2.12 & 3.85 & 2.40 & 0.27  \\\cline{1-8}
    			PDSNet\cite{tulyakov2018practical} & 2.29 & 4.05 & 2.58 & 2.09 & 3.69 & 2.36 & 0.50  \\\cline{1-8}
    		    PSMNet\cite{chang2018pyramid} & 1.86 & 4.62 & 2.32 & 1.71 & 4.31 & 2.14 & 0.41  \\\cline{1-8}
    		    SegStereo\cite{yang2018segstereo} & 1.88 & 4.07 & \textbf{2.25} & 1.76 & 3.70 & 2.08 & 0.60  \\\cline{1-8}
    		    Ours & \textbf{1.72} & 4.95 & 2.26 & \textbf{1.57} & 4.58 & \textbf{2.06} & 0.23  \\\cline{1-8}
    		\end{tabular}
    	\caption{Comparison of disparity estimation from StereoDRNet with state of the art published methods on KITTI 2015 dataset.}
    	\label{Table:KITT2015}
    \end{table}
    
    \begin{table}[h]
    	\centering
    	\footnotesize \setlength\tabcolsep{3.0pt}\renewcommand{\arraystretch}{1.2}
    		\begin{tabular}{|c|c|c|c|c|c|c|c|c|}
    			\hline\multicolumn{1}{|c|}{\multirow{2}{*}{Method}} & \multicolumn{4}{|c|}{\multirow{1}{*}{All}} & \multicolumn{4}{|c|}{\multirow{1}{*}{Noc}}\\\cline{2-9}
    			
    			\multicolumn{1}{|c|}{} & \multicolumn{1}{|c|}{1px} & \multicolumn{1}{|c|}{2px} & \multicolumn{1}{|c|}{4px} & \multicolumn{1}{|c|}{RMSE} & \multicolumn{1}{|c|}{1px} & \multicolumn{1}{|c|}{2px} & \multicolumn{1}{|c|}{4px} & \multicolumn{1}{|c|}{RMSE}  \\\cline{1-9}
    			
    			PSMNet\cite{chang2018pyramid} & 5.41 & 1.31 & 0.54 & 0.75 & 5.02 & 1.09 & 0.41 & 0.66  \\\cline{1-9}
    			iResNet\cite{liang2018learning} & 4.04 & 1.20 & 0.34 & 0.59 & 3.68 & 1.00 & 0.25 & 0.51 \\\cline{1-9}
    			DN-CSS\cite{ilg2018occlusions} & \textbf{3.00} & 0.96 & 0.34 & 0.56 & \textbf{2.69} & \textbf{0.77} & 0.26 & \textbf{0.48}  \\\cline{1-9}
    		    Ours & 4.84 & \textbf{0.96} & \textbf{0.30} & \textbf{0.55} & 4.46 & 0.83 & \textbf{0.24} & 0.50 \\\cline{1-9}
    		\end{tabular}
    	\caption{Comparison of disparity estimation from StereoDRNet with state of the art published methods on ETH 3D dataset. }
    	\label{Table:ETH3D}
    \end{table}

    \vspace{-0.55cm}
    \subsection{KITTI Datasets}
    
    We evaluated our method on both KITTI 2015 and KITTI 2012 datasets. These data sets contain stereo pairs with semi-dense depth images acquired using a LIDAR sensor that can be used for training. The KITTI 2012 dataset contains 194 training and 193 test stereo image pairs from static outdoor scenes. The KITTI 2015 dataset contains 200 training and 200 test stereo image pairs from both static and dynamic outdoor scenes. 
    
    \noindent\textbf{Training and ablation study:} Since KITTI data sets contain only limited amount of training data, we fine tuned our model on the SceneFlow dataset. In our training we used 80\% stereo pairs for training and 20\% stereo pairs for evaluation. We demonstrate the ablation study of our proposed method on KITTI 2015 dataset Table~\ref{Table:AbalationNetwork}. Note how our proposed dilated residual architecture and the use of Vortex pooling for feature extraction consistently improve the results. We did not achieve significant gains by doing refinement on KITTI datasets as these datasets only contain labeled depth for sparse pixels. Our refinement procedure improves disparity predictions using view consistency checks and sparsity in ground truth data affected the training procedure. We demonstrate that data sets with denser training data enabled the training and fine-tuning of our refinement model. 
    
    \noindent\textbf{Results:}  We evaluated our Dilated residual network without filtering on both these datasets and achieved state of the art results on KITTI 2012  Table~\ref{Table:KITT2012} and comparable results with best published method on KITTI 2015 Table \ref{Table:KITT2015}. On KITTI 2015 dataset the three columns “D1-bg”, “D1-fg” and “D1-all” mean that the pixels in the background, foreground, and all areas, respectively, were considered in the estimation of errors. We perform consistently well in “D1-bg” meaning background areas, we achieve comparable results with state of art method in all pixels and better results in non-occluded regions. On KITTI 2012 dataset "Noc" means non occluded regions and "All" mean all regions. Notice, that we perform comparable against SegStereo~\cite{yang2018segstereo} on KITTI 2015 but way better in KITTI 2012 dataset.
    
    \subsection{ETH3D Dataset} We again used our pre-trained network trained on Sceneflow dataset and fine-tuned it on the training set provided in the dataset. ETH dataset contains challenging scenes of both outside and indoor environment. According to our Table ~\ref{Table:ETH3D} we perform best on almost half of the evaluation metrics, our major competitor in this evaluation was DN-CSS ~\cite{ilg2018occlusions}. Although, we observe that this method did not perform well on KITTI 2015 data set Table ~\ref{Table:KITT2015}. Notice, as this data set contained dense training disparity maps of both stereo views we were able to train and evaluate our refinement network on this data set.
    
	\subsection{Indoor Scene Reconstruction}
	\label{sec:SceneReconstruction}
	We use the scanning rig used in recent work \cite{whelan2018reconstructing} for preparing ground truth dataset for supervised learning of depth and added one more RGB camera to the rig to obtain a stereo image pair. We kept the baseline of the stereo pair to be about 10cm. We trained our StereoDRNet network on SceneFlow as described in section 4.1 and then fine tuned the pre-trained network on 250 stereo pairs collected in the indoor area by our scanning rig. We observed that the network to quickly adapted to our stereo rig with a minimal amount of fine-tuning. 
	
	For preparing ground truth depth we found rendered depth from complete scene reconstruction to be a better estimate than the live sensor depth which usually suffers from occlusions and depth uncertainties. Truncated signed distance function (TSDF) was used to fuse live depth maps into a scene as described in \cite{newcombe2011kinectfusion}.
	
	\begin{figure}[h]
	\centering
	\includegraphics[width=1.0\linewidth]{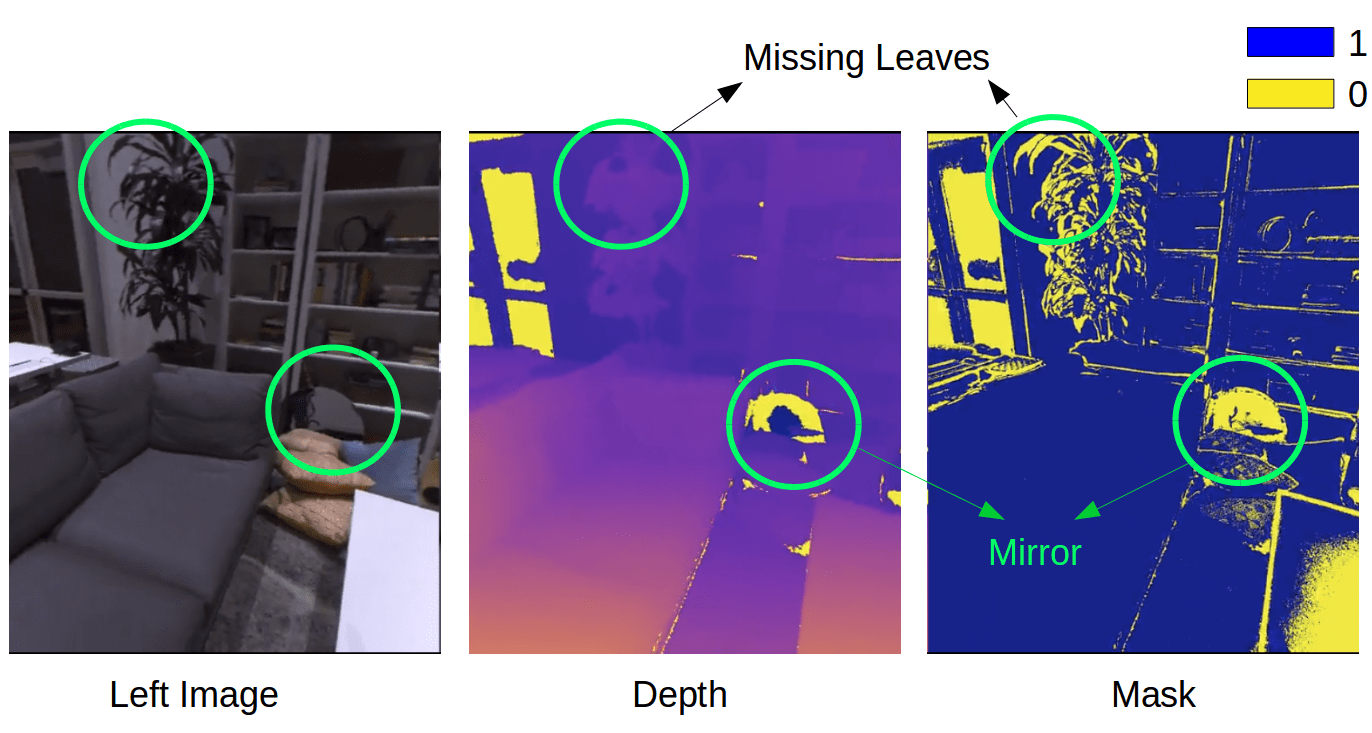}
		\caption{We show a training example with the left image, ground truth depth and the exclusion mask. Note that the glass, mirrors and the sharp corners of the table are excluded from training as indicated by the yellow pixels in the occlusion mask. Note, that this example was not part of our actual training set.} 
		\label{fig:IndoorTraining}
	\end{figure}
	
   \begin{figure}[h]
   \centering
   		\includegraphics[width=1.0\linewidth]{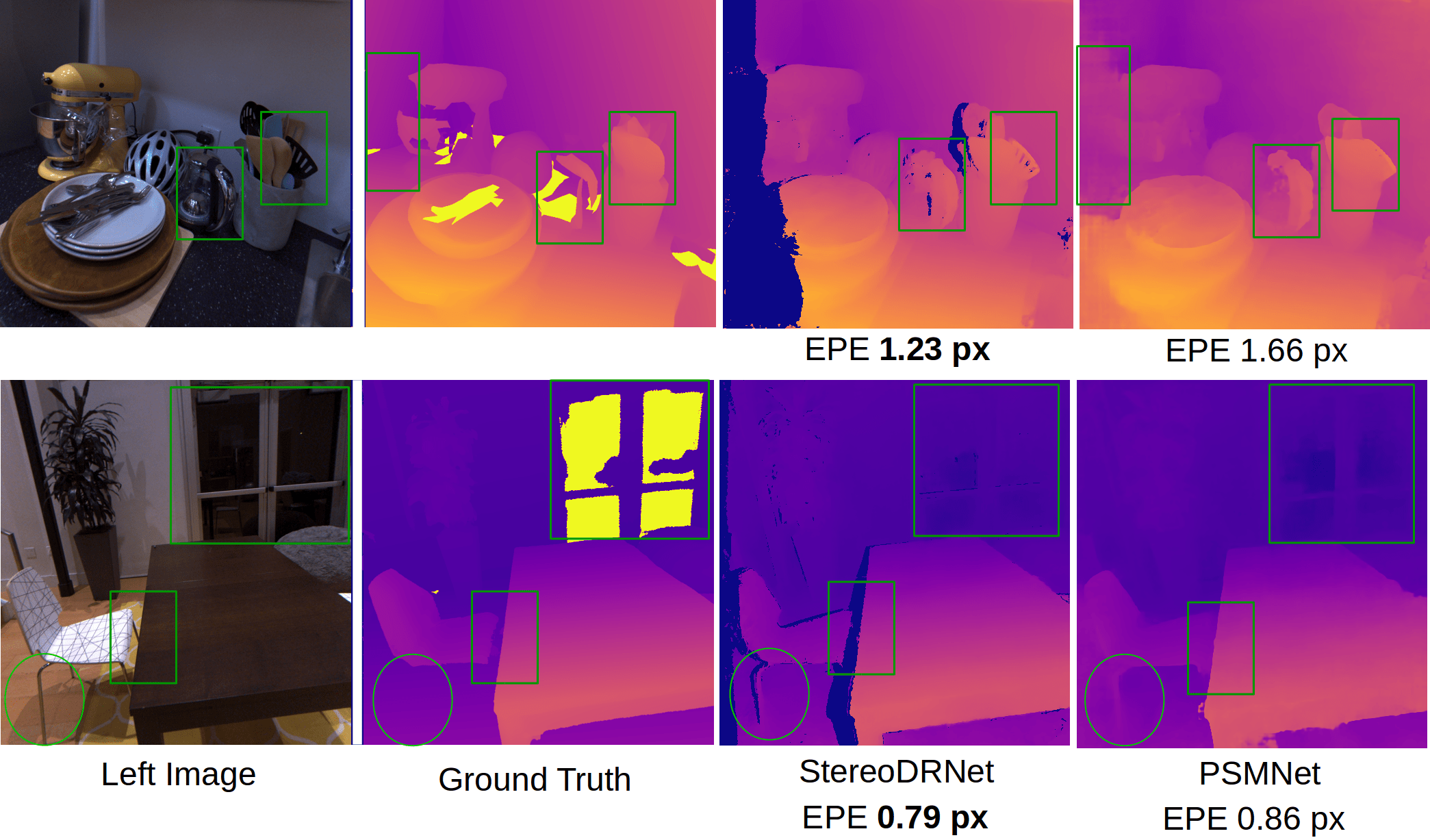}
   	\caption{ This figure demonstrates that our StereoDRNet network produces better predictions on thin reflective legs of the chair and some portions of the glass. We used occlusion mask predicted by our network to clip occluding regions. Yellow region in the ground truth are the regions that belong to our proposed exclusion mask. } 
   	\label{fig:IndoorResult}
   \end{figure}
   
   \begin{figure}[h]
     \centering
   		\includegraphics[width=1.0\linewidth]{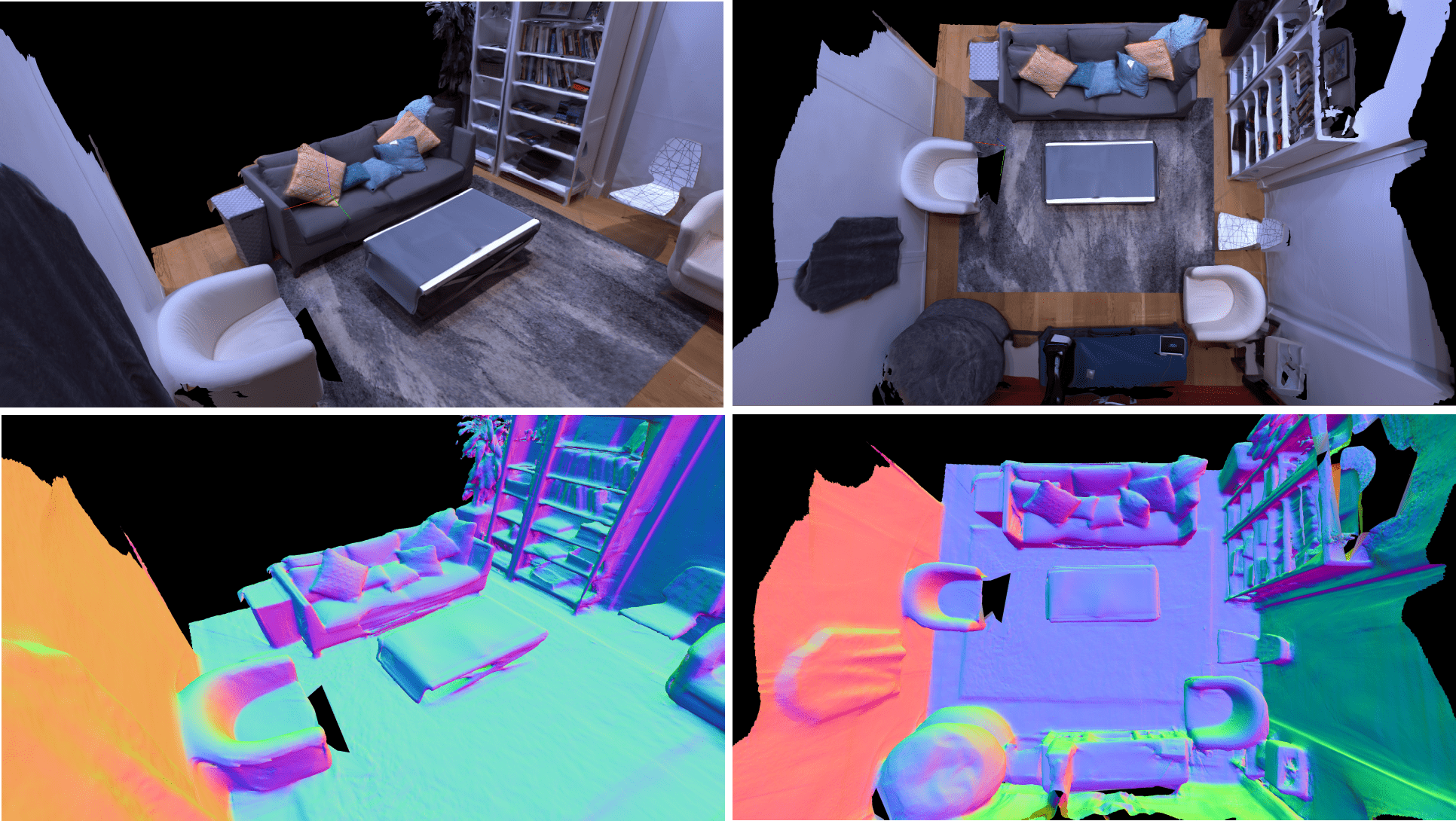}
   	\caption{ This figure demonstrates 3D reconstruction of a living room in an apartment prepared by TSDF fusion of the predicted depth maps from our system. We visualize two views of the textured mesh and surface normals in top and bottom rows respectively.} 
   	\label{fig:IndoorReconstruction}
   \end{figure}
   
    \begin{figure*}[h]
    \centering
    \includegraphics[width=0.85\textwidth,clip,trim=0 0 0 420]{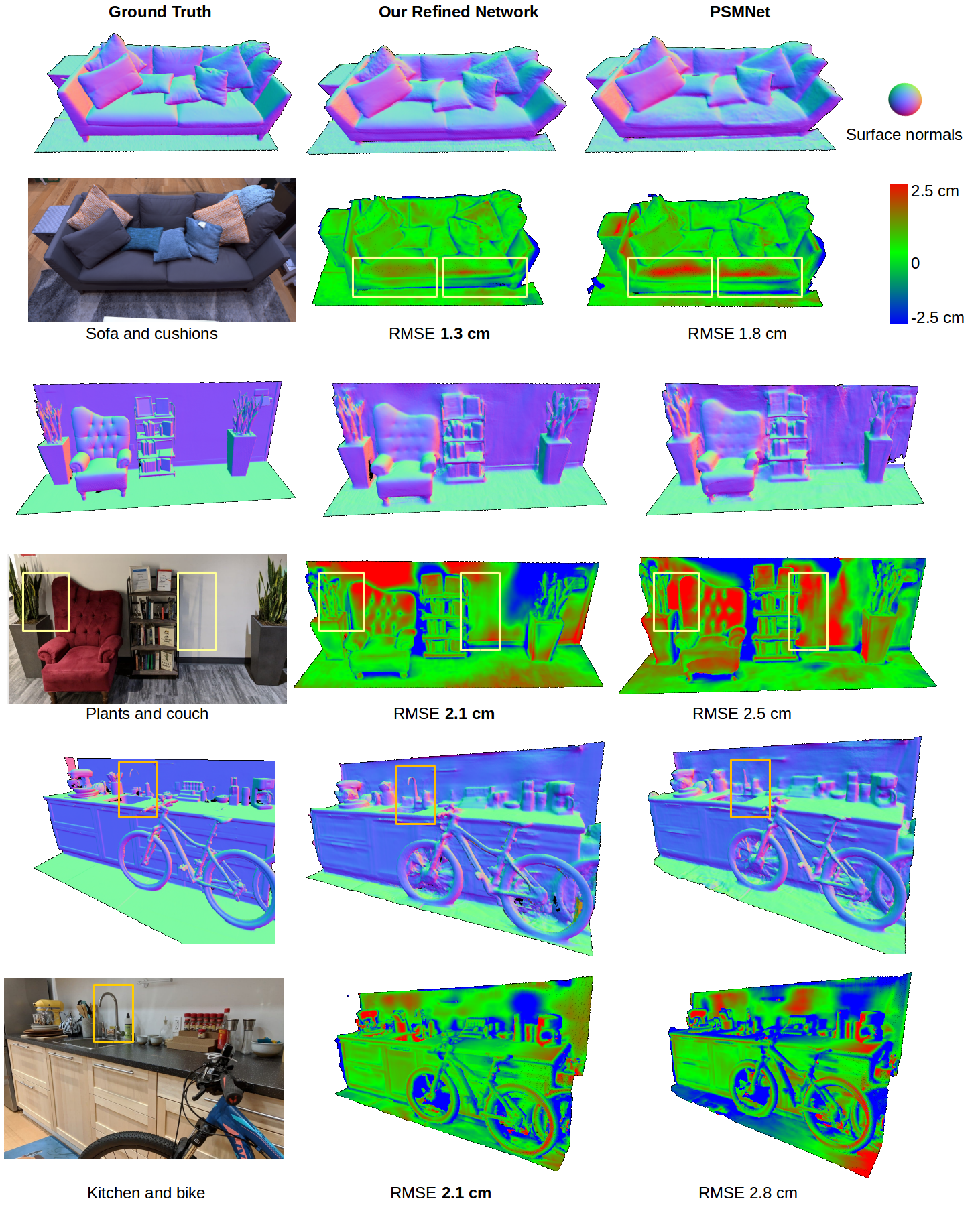}
   	\caption{Comparison of 3D reconstruction using fusion of depth maps from our StereoDRNet network (middle), PSMNet~\cite{chang2018pyramid} (right) and depth maps from the structured light system (left) described in \cite{whelan2018reconstructing} (termed Ground Truth). We report and visualize point-to-plane distance RMS error on the reconstructed meshes with respect to the ground truth mesh. Dark yellow boxes represent the regions where our reconstruction yields details that the structured light sensor or PSMNet were not able to capture. Light yellow boxes represent regions where StereoDRNet outperforms PSMNet.} 
   	\label{fig:ReconstructionComparison}
   \end{figure*}

The infrared-structure light depth sensors are known to be unresponsive to dark and highly reflective surfaces. Moreover, the quality of TSDF fusion is limited to the resolution of the voxel size. Hence we expect the reconstructions to be overly smooth in some areas such as table corners or sharp edges of plant leaves. In order to avoid contaminating our training data with false depth estimation, we use a simple photometric error threshold to mask out the pixels from training where the textured model projection color disagrees with the real images. We show one such example in Fig.~\ref{fig:IndoorTraining} where glass, mirrors and the sharp corners of the table are excluded from training. Although, the system from Whelan~et~al.~\cite{whelan2018reconstructing} can obtain ground truth planes of mirrors and glass we avoid depth supervision on them in this work as it is beyond the scope of a stereo matching procedure to obtain depth on reflectors.\\
We demonstrate visualizations of the depth predictions from the stereo pair in Fig.~\ref{fig:IndoorResult}. Notice, our prediction is able to recover sharp corners of the table, thin reflective legs of the chair and several thin structures in kitchen dataset as a result of filtering process used in training. It is interesting to see that we recover the top part of the glass correctly but not the bottom part of the glass which suffers from reflections. The stereo matching model simply treats reflectors as windows in presence of reflections.\\
\noindent\textbf{Results and evaluations:} We demonstrate visualizations of full 3D reconstruction of a living room in an apartment prepared by TSDF fusion of the predicted depth maps from our system in Fig.~\ref{fig:IndoorReconstruction}. For evaluation study we prepared three small data sets that we refer as ``Sofa and cushions'' demonstrated in Fig.~\ref{fig:teaser}, ``Plants and couch'' and ``Kitchen and bike'' demonstrated in Fig.~\ref{fig:ReconstructionComparison}. We report point-to-plane root mean squared error (RMSE) of the reconstructed 3D meshes from fusion of depth maps obtained from PSMNet~\cite{chang2018pyramid} and our refined network. We obtain a RMSE of 1.3 cm on the simpler ``Sofa and cushions'' dataset. Note that our method captured high frequency structural details on the cushions which were not captured by PSMNet or the structured light sensor. ``Plants and couch'' represents a more difficult scene as it contained a directed light source casting shadows. For this dataset StereoDRNet obtained 2.1 cm RMSE whereas PSMNet obtained 2.5 cm RMSE. Notice, that our reconstruction is not only cleaner but produces minimal errors in the shadowed areas (shadows cast by book shelf and left plant). ``Kitchen and bike'' dataset cluttered and contains reflective objects making it the hardest dataset. While our system still achieved 2.1 cm RMSE, the performance of PSMNet degraded to 2.8 cm RMSE. Notice, that our reconstruction contains the faucet (highlighted by yellow box) in contrast to the structured light sensor and PSMNet reconstructions. For all evaluations we used exactly the same training dataset for fine-tuning our StereoDRNet and PSMNet.\\    
%    We used DSO system to provide camera tracking while fusing depth maps into the volume, marching cubes is used to extract mesh from TSDF and mesh is later on textured by left camera stream. The whole apartment was scanned by handheld scan of about 5 minutes. 
\vspace{-0.75cm}
\section{Conclusion}
\vspace{-0.15cm}
 Depth estimation from passive stereo images is a challenging task. Systems from related work suffer in regions with homogeneous texture or surfaces with shadows and specular reflections. Our proposed network architecture uses global spatial pooling and dilated residual cost filtering techniques to approximate the underlying geometry even in above mentioned challenging scenarios. Furthermore, our refinement network produces geometrically consistent disparity maps with the help of occlusion and view consistency cues. The use of perfect synthetic data and careful filtering of real training data enabled us to recover thin structures and sharp object boundaries. Finally, we demonstrate that our passive stereo system, when used for building 3D scene reconstructions in challenging indoor scenes, approaches the quality of state-of-the-art structured light systems \cite{whelan2018reconstructing}. 
%  This opens up the ability to capture in outdoor environments where active depth sensing methods can fail. With the advent of mobile phones, drones and other robotic devices with stereo cameras, the potential to enable democratization of high quality model capture is enticingly close. 
 
% (Julian) I am not a fan of putting to much future work in there - it kind of just shows what you didnt get to doing
%In future we plan to utilize semantic scene information to guide our disparity estimation system. We also plan to extend our existing architecture to exploit small camera motion to derive better object and occlusion boundaries.

\section*{Supplementary}
\setcounter{section}{0}
\renewcommand\thesection{\Alph{section}}
\section{Overview}
	In this supplementary material, we provide additional details of the training and evaluation procedure of our indoor scene reconstruction experiments. We also provide in depth detail of our proposed network architecture and show the effect of the proposed refinement procedure on the reconstruction quality. We share the results of the ablation study on the dilated convolutions used in our cost filtering approach and visualize the comparison of the disparity predictions from our system with state of art methods on KITTI and ETH3D benchmarks.
	
	\begin{figure}[h]
	\centering
			\includegraphics[width=0.9\linewidth]{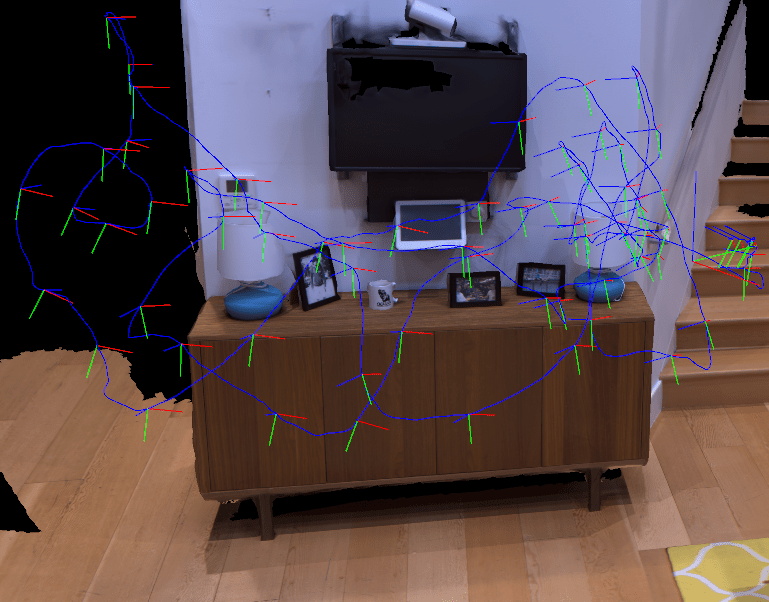}
	  	\caption{This figure shows the training scene used for all our indoor scene reconstruction experiments. We used about 200 stereo views rendered by OpenGL from poses along the camera trajectory visualized by the blue curve. The 3D reconstruction was built using the method described in \cite{whelan2018reconstructing}.} 
	   	\label{fig:Training}
	\end{figure} 
	
	\section{3D Reconstruction Experiments}
	For all 3D reconstruction experiments and evaluations we used a set of about 200 stereo views shown in Fig.~\ref{fig:Training} to fine tune the SceneFlow~\cite{mayer2016large}-pre-trained networks. 
	
\begin{figure*}[h]
	\centering
			\includegraphics[width=0.9\textwidth]{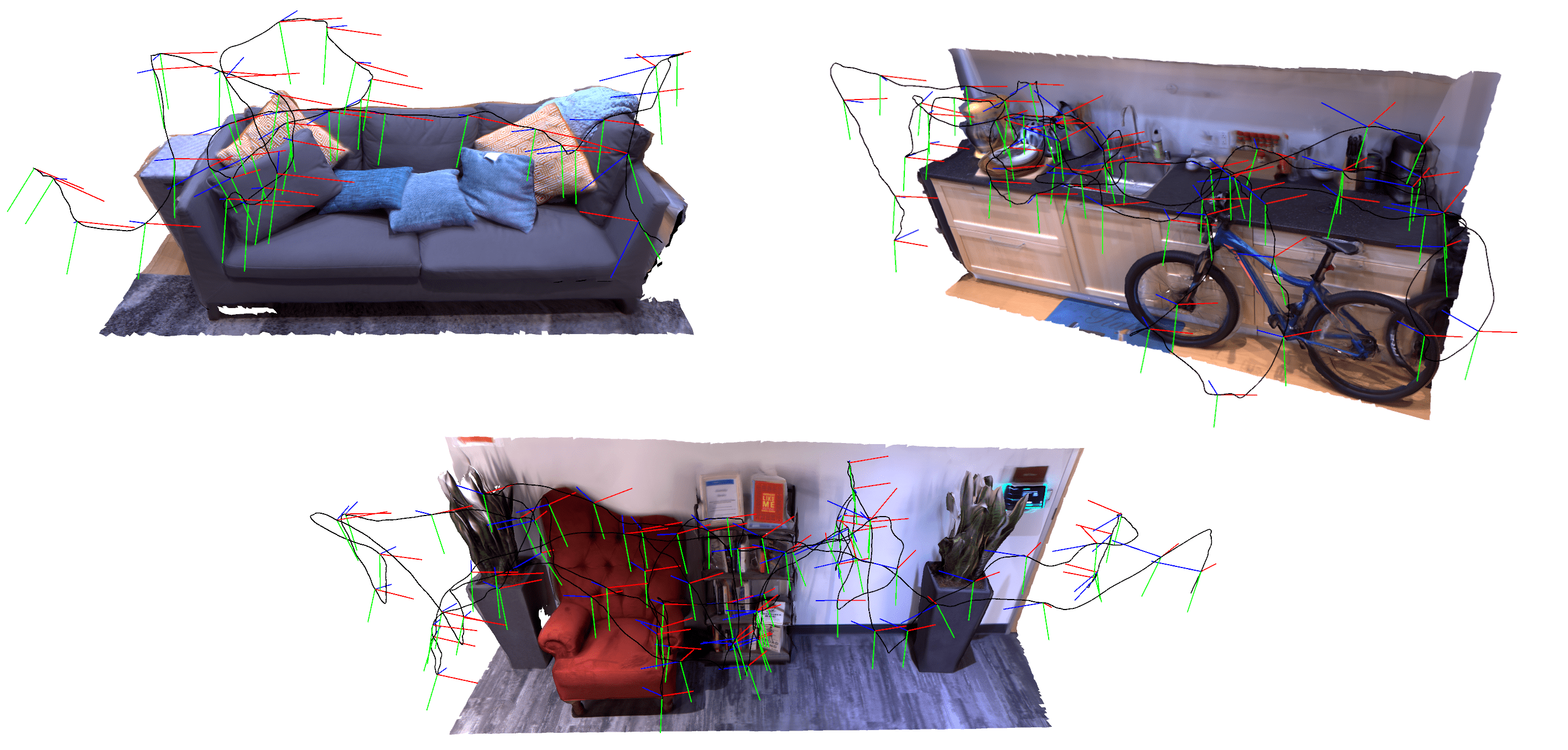}
	  	\caption{This figure shows the textured 3D reconstructions of "Sofa and cushions", "Plants and couch" and "kitchen and bike" scenes developed using KinectFusion~\cite{newcombe2011kinectfusion,whelan2018reconstructing} of depth maps generated form StereoSDRNet with refinement. We visualize the camera trajectory, from which the stereo images were taken, via a black curve. Note that for clarity we visualize every 30th frame used by the fusion system.} 
	   	\label{fig:EvaluationPoses}
	\end{figure*}

    We show the textured 3D reconstructions of our indoor scene dataset in Fig.~\ref{fig:EvaluationPoses}. Note that we used KinectFusion~\cite{newcombe2011kinectfusion} to fuse the depth maps into 3D spatial maps. We did not use any structure-from-motion (SfM) or external localization method for estimating camera trajectories. Hence, the camera views visualized in Fig.~\ref{fig:EvaluationPoses} are the output of the ICP (iterative closest point) procedure used by the KinectFusion~\cite{newcombe2011kinectfusion} system. We used manual adjustment followed by ICP to align the 3D reconstructions wherever necessary for our evaluations.
    
    \section{Network Details}
    We provide the network architecture of StereoDRNet in Table.~\ref{Table:Network}. We borrowed ideas on extracting robust local image features from PSMNet~\cite{chang2018pyramid}. As described in the paper, we use Vortex Pooling~\cite{xie2018vortex} for extracting global scene context. In our experiments we found dilation rates 3, 5 and 15 and average grids of size $3\times 3$, $5\times 5$ and $15 \times 15$ to improve performance more in disparity predictions than the one proposed in the original work for semantic segmentation. 
    
    \begin{table}[h]
    	\centering
    	\footnotesize \setlength\tabcolsep{3.2pt} \renewcommand{\arraystretch}{1.2}
    		\begin{tabular}{|c|c|c|c|c|}
    		    \hline
    		    \multicolumn{4}{|c|}{3D Dilation in Cost Filtering} & \multicolumn{1}{|c|}{SceneFlow}\\\cline{1-4}
    		    
    			rate = 1&rate = 2&rate = 4&rate = 8&EPE\\\hline
    			\checkmark & & & & 1.13\\\hline
    			\checkmark & \checkmark & & & 1.03\\\hline
    			\checkmark & \checkmark & \checkmark & & \textbf{0.98}\\\hline
    			\checkmark & \checkmark & \checkmark & \checkmark & 1.01\\
		        \hline
    		
    		\end{tabular}
    	\caption{Ablation study of dilated convolution rates used in the proposed dilated cost filtering scheme. Note that we used StereoDRNet without refinement in this study.}
    	\label{Table:DilationAbalation}
    \end{table}
    
    In order to show the effectiveness of the proposed dilated convolutions in cost filtering, we conduct an ablation study in Table.~\ref{Table:DilationAbalation} on the SceneFlow~\cite{mayer2016large} dataset. We observed that increasing dilation rates improved the quality of predictions. Dilation rates above 4 did not provide any significant gains.
    
    \begin{table}[ht]
    	\centering
    	\footnotesize \setlength\tabcolsep{3.0pt}\renewcommand{\arraystretch}{1.2}
    		\begin{tabular}{|l|l|l|}
    			\hline Index  & Layer Description & Output \\
    			\hline
    			1 & Warp($I_R$,$\mathbf{d_L^3}$) - $I_L$ & H x W x 3 \\ 
    			2 & concat 1, $I_L$ & H x W x 6 \\ 
    			3 & Warp($\mathbf{d_R^3}$, $\mathbf{d_L^3}$) - $\mathbf{d_L^3}$ & H x W x 1 \\
    			4 & concat 3, $\mathbf{d_L^3}$ & H x W x 2 \\
    			5 & 3x3 conv on 2, 16 features & H x W x 16\\
    			6 & 3x3 conv on 4, 16 features & H x W x 16\\
    			7 & concat 5,6 $I_L$ & H x W x 32 \\
    			\multirow{2}{*}{8-13} & (3x3 conv, residual block) x 6,  & \multirow{2}{*}{H x W x 32} \\
    			& dil rate 1,2,4,8,1,1&\\
    			14 & 3x3 conv, 2 features as 14(a) and 14(b)  & H x W x 2 \\
    			15 & $\mathbf{d^r}$: 14(a) + $\mathbf{d_L^3}$ & H x W  \\
    			16 & \textbf{O}: sigmoid on 14(b) & H x W  \\
    			\hline
    		\end{tabular}
    	\caption{Refinement network for StereoDRNet. $\mathbf{d^r}$ and \textbf{O} represent refined disparity and occlusion probability respectively.   }
    	\label{Table:RefinementNetwork}
    \end{table}
    \begin{table}[ht]
    	\centering
    	\footnotesize \setlength\tabcolsep{3.0pt} \renewcommand{\arraystretch}{1.15}
    		\begin{tabular}{|l|l|l|}
    			\hline Index & Layer Description & Output \\
    			\hline
    			1 & Input Image & H x W x 3 \\
    			\hline
    			\multicolumn{3}{|c|}{Local feature extraction}\\
    			\hline
    			2 & 3x3 conv, 32 features, stride 2 & H/2 x W/2 x 32\\
    			3-4 & (3x3 conv, 32 features) x 2 & H/2 x W/2 x 32\\
    			5-7 & (3x3 conv, 32 features, res block) x 3 & H/2 x W/2 x 32\\
    			8 & 3x3 conv, 32 features, stride 2 & H/4 x W/4 x 32\\
    			9-22 & (3x3 conv, 64 features, res block) x 15 & H/4 x W/4 x 64\\
    			23-28 & (3x3 conv, 128 features, res block) x 6 & H/4 x W/4 x 128\\
    			\hline
    			\multicolumn{3}{|c|}{Spatial Pooling}\\
    			\hline
    			29 & Global Avg Pool on 28, bi-linear interp & H/4 x W/4 x 128\\
    			30 & Avg Pool 3x3 on 28, conv 3x3, dil rate 3 & H/4 x W/4 x 128\\
    		    31 & Avg Pool 5x5 on 28, conv 3x3, dil rate 5 & H/4 x W/4 x 128\\
    		    \multirow{2}{*}{32} & Avg Pool 15x15 on 28, conv 3x3, & \multirow{2}{*}{H/4 x W/4 x 128}\\
    		    &dil rate 15& \\
    		    33 & Concat 22, 28, 29, 30, 31 and 32 & H/4 x W/4 x 704\\ 
    		    34 & 3x3 conv, 128 features & H/4 x W/4 x 128\\
    		    \multirow{2}{*}{35} & 1 x 1 conv, 32 features without BN& \multirow{2}{*}{H/4 x W/4 x 32}\\
    		    &and ReLU&\\
    		    \hline
    			\multicolumn{3}{|c|}{Cost Volume}\\
    			\hline
    			\multirow{2}{*}{36} & Subtract left 35 from right 35& \multirow{2}{*}{D/4 x H/4 x W/4 x 64}\\
    			&with D/4 shifts,vice versa&\\
    			\hline
    			\multicolumn{3}{|c|}{Cost Filtering}\\
    			\hline
    			37-38 & (3x3x3 conv, 32 features) x 2 & D/4 x H/4 x W/4 x 32\\ 
    			39 & 3x3x3 conv, 32 features, stride 2 & D/8 x H/8 x W/8 x 32\\
    			40 & 3x3x3 conv, 32 features & D/8 x H/8 x W/8 x 32\\
    			41 & 3x3x3 conv on 39, 32 features & D/8 x H/8 x W/8 x 32\\
    			42 & 3x3x3 conv on 39, 32 features, dil rate 2 & D/8 x H/8 x W/8 x 32\\
    			43 & 3x3x3 conv on 39, 32 features, dil rate 4 & D/8 x H/8 x W/8 x 32\\
    			\multirow{2}{*}{44} & 3x3x3 conv on concat(41,42,43), & \multirow{2}{*}{D/8 x H/8 x W/8 x 32}\\
    			&32 features&\\
    			45 & 3x3x3 deconv, 32 features, stride 2 & D/4 x H/4 x W/4 x 32\\
    			46 & \textbf{Pred1}: 3x3x3 conv on 45 + 38  & D/4 x H/4 x W/4 x 2 \\
    			47 & 3x3x3 conv on 45, 32 features, stride 2 & D/8 x H/8 x W/8 x 32\\
    			48 & 3x3x3 conv + 40, 32 features & D/8 x H/8 x W/8 x 32\\
    			49 & 3x3x3 conv on 48, 32 features & D/8 x H/8 x W/8 x 32\\
    			50 & 3x3x3 conv on 48, 32 features, dil rate 2 & D/8 x H/8 x W/8 x 32\\
    			51 & 3x3x3 conv on 48, 32 features, dil rate 4 & D/8 x H/8 x W/8 x 32\\
    			\multirow{2}{*}{52} & 3x3x3 conv on concat(49,50,51), & \multirow{2}{*}{D/8 x H/8 x W/8 x 32}\\
    			&32 features&\\
    			53 & 3x3x3 deconv, 32 features, stride 2 & D/4 x H/4 x W/4 x 32\\
    			54 & \textbf{Pred2}: 3x3x3 conv on 53 + 38  & D/4 x H/4 x W/4 x 2 \\
    			55 & 3x3x3 conv on 53, 32 features, stride 2 & D/8 x H/8 x W/8 x 32\\
    			56 & 3x3x3 conv + 48, 32 features & D/8 x H/8 x W/8 x 32\\
    			57 & 3x3x3 conv on 56, 32 features & D/8 x H/8 x W/8 x 32\\
    			58 & 3x3x3 conv on 56, 32 features, dil rate 2 & D/8 x H/8 x W/8 x 32\\
    			59 & 3x3x3 conv on 56, 32 features, dil rate 4 & D/8 x H/8 x W/8 x 32\\
    			\multirow{2}{*}{60} & 3x3x3 conv on concat(57,58,59), & \multirow{2}{*}{D/8 x H/8 x W/8 x 32}\\
    			&32 features&\\
    			61 & 3x3x3 deconv, 32 features, stride 2 & D/4 x H/4 x W/4 x 32\\
    			62 & \textbf{Pred3}: 3x3x3 conv on 61 + 38  & D/4 x H/4 x W/4 x 2 \\
    			\hline
    			\multicolumn{3}{|c|}{Disparity Regression}\\
    			\hline
    			63 & Bi-linear interp of \textbf{Pred1}, \textbf{Pred2}, \textbf{Pred3} &  D x H x W x 2 \\
    			64 & SoftArg Max of 63 to get $\mathbf{d^1}$, $\mathbf{d^2}$, $\mathbf{d^3}$ & H x W x 2 \\
    			\hline
    		\end{tabular}
    	\caption{Full StereoDRNet architecture. Note that when used without refinement, StereoDRNet just outputs $\mathbf{d^1}$, $\mathbf{d^2}$ and $\mathbf{d^3}$ for the left view. }
    	\label{Table:Network}
    \end{table}
    
     The proposed refinement network described in  Table.~\ref{Table:RefinementNetwork} is inspired by the refinement procedures proposed in CRL~\cite{pang2017cascade},~ iResNet~\cite{liang2018learning},~ StereoNet~\cite{khamis2018stereonet}, and ActiveStereoNet~\cite{zhang2018activestereonet}. We adopted the basic architecture for refinement as described in StereoNet~\cite{khamis2018stereonet} with dilated residual blocks~\cite{yu2017dilated} to increase the receptive field of filtering without compromising resolution. This technique was also adopted in recent work on optical flow prediction Pwc-net~\cite{sun2018pwc}. We experienced additional gains when using the photometric error $E_p$ and geometric error maps $E_g$ as inputs and co-training of occlusion maps. Such enhancements in the refinement procedure has never been proposed to the best of our knowledge. 
	
	\section{Effect of Refinement}
	\label{sec:refinementeffect}
\begin{figure*}[h]
	\centering
			\includegraphics[width=\textwidth]{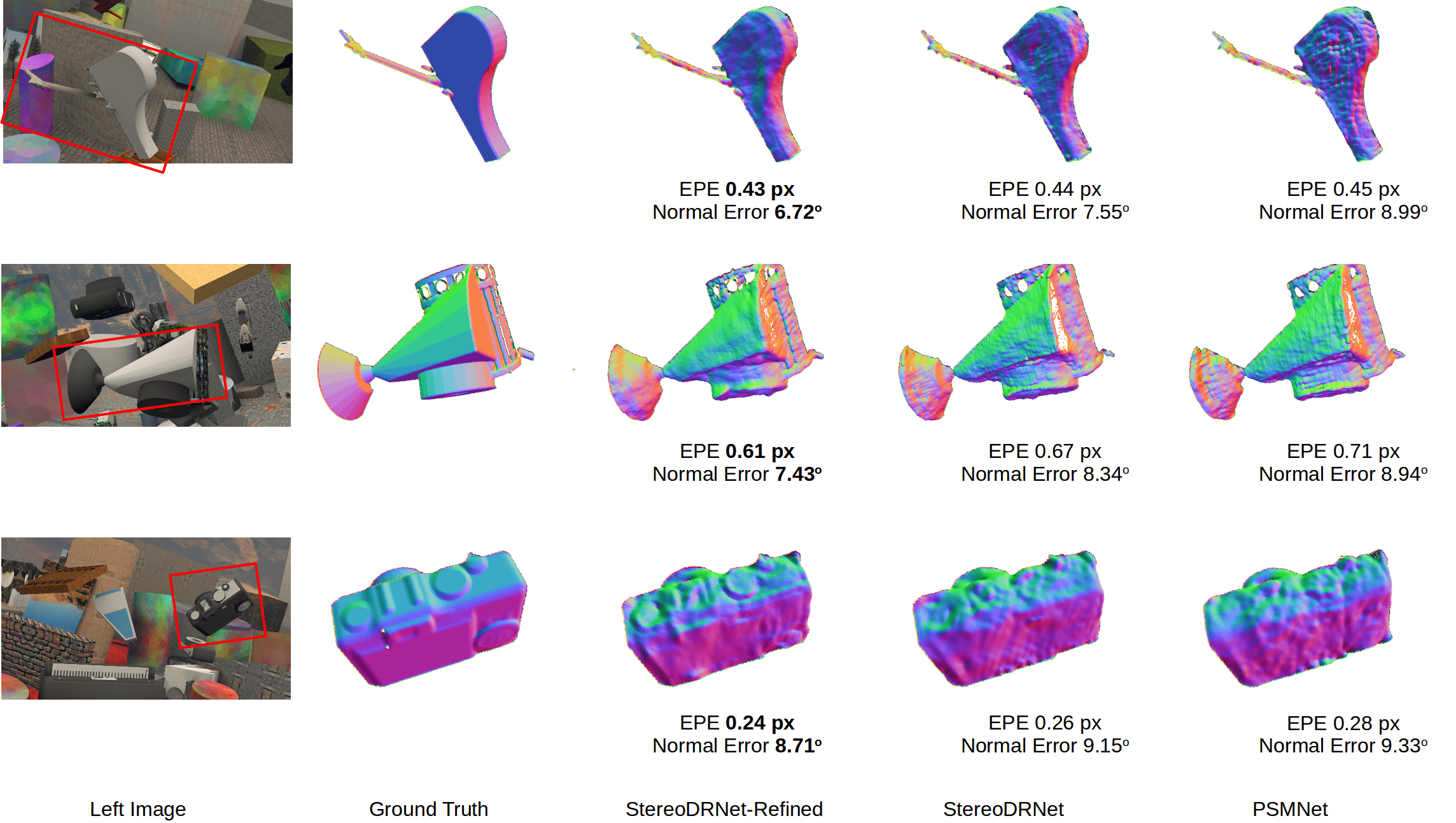}
	  	\caption{ This figure demonstrates the surface normal visualizations of some objects (labeled with red boxes) reconstructed using \textbf{single} disparity map from SceneFlow dataset. We report EPE in disparity space and surface normal error in degrees. Notice, our refinement network improves the overall structure of the objects and makes them geometrically consistent. } 
	   	\label{fig:NormalSceneFlow}
	\end{figure*}

\begin{figure*}[h]
	\centering
			\includegraphics[width=\textwidth]{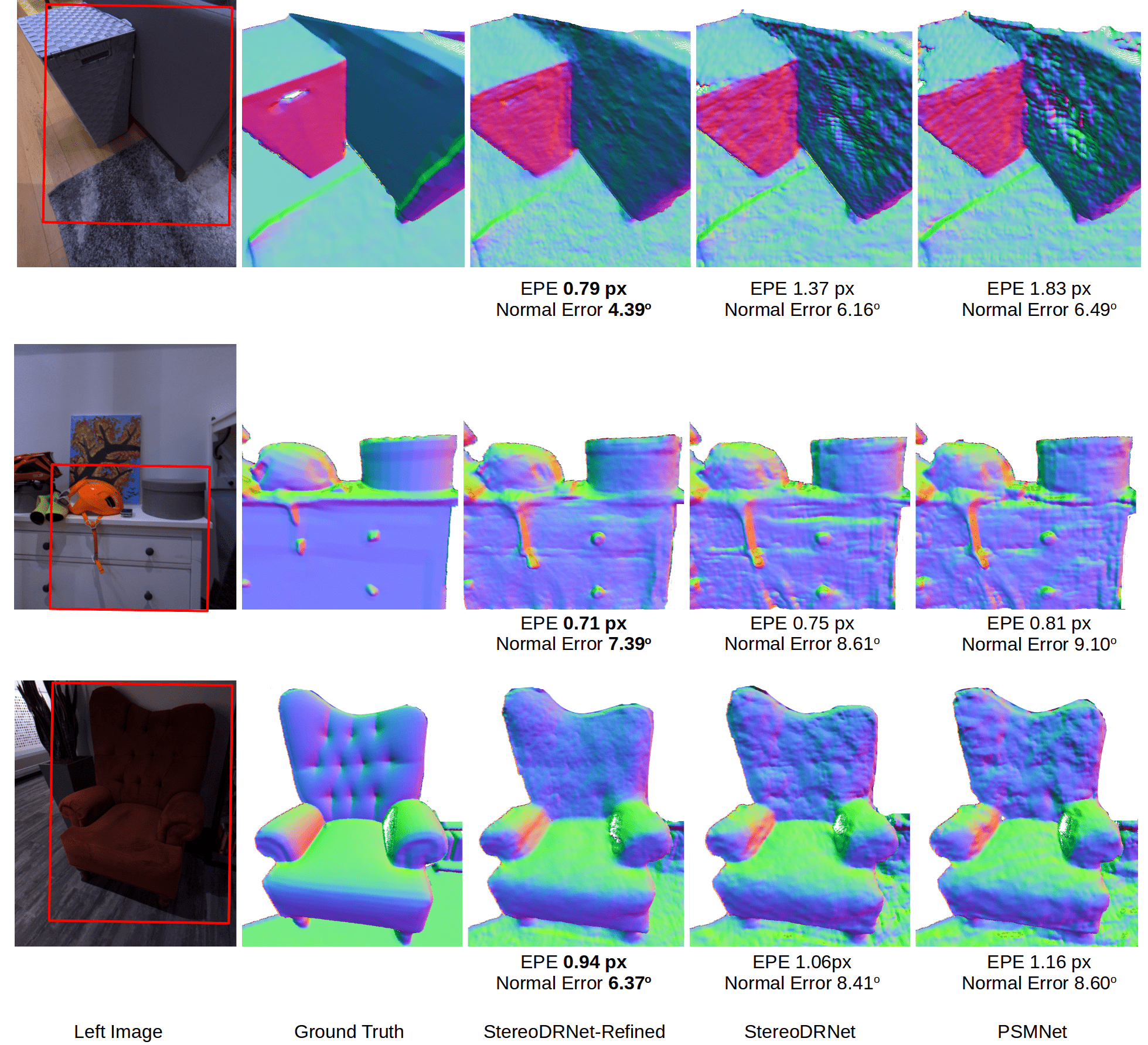}
	  	\caption{ This figure shows the surface normal visualizations of some objects (labeled with red boxes) reconstructed using a \textbf{single} disparity map from our real dataset. We report EPE in disparity space and surface normal error in degrees. Notice that our refinement network improves the overall structure of the objects and makes them geometrically consistent.} 
	   	\label{fig:NormalReal}
	\end{figure*}

	Our refinement procedure not only improves the overall disparity error but also makes the prediction geometrically consistent. We calculate surface normal maps from disparity/depth maps using the approach described in KinectFusion~\cite{newcombe2011kinectfusion}. We use a surface normal error metric to measure consistency in the disparity predictions (first order derivative).  Figures~\ref{fig:NormalSceneFlow}~and~\ref{fig:NormalReal} visualize how our refinement procedure improves the overall structure of objects. In some cases such as in the first comparison in Fig.~\ref{fig:NormalSceneFlow} we observe little improvement in disparity prediction but large improvement in surface normals. Figure~\ref{fig:NormalReal} demonstrates real scene disparity and derived surface normal predictions and proves that our refinement procedure works well on real world data in presence of shadows and dark lighting conditions. Dense 3D reconstruction methods such as KinectFusion~\cite{newcombe2011kinectfusion} use surface normals to calculate fusion parameters and confidence weights, hence it is important to predict geometrically consistent disparity or normal maps for high quality 3D reconstruction.

	\vspace*{-5.5in}
	\begin{figure*}[ht!]
	\centering
			\includegraphics[width=\textwidth]{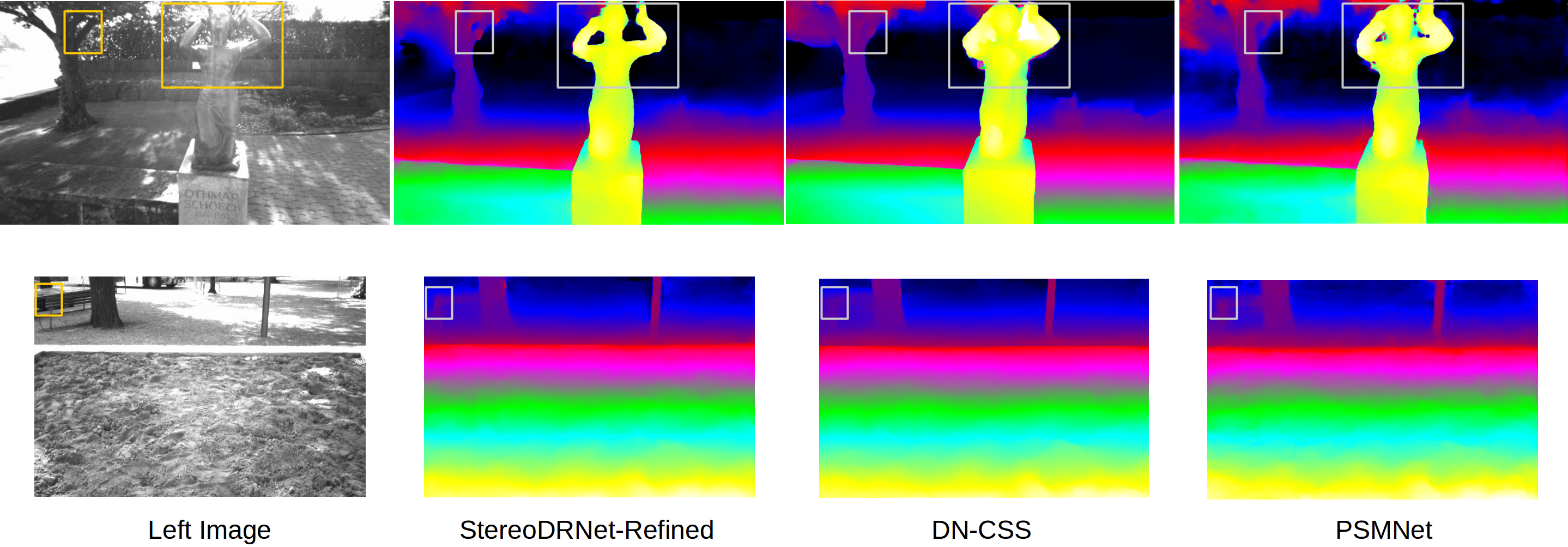}
	  	\caption{ This figure shows the disparity estimation results of our refined network, PSMNet~\cite{chang2018pyramid} and DN-CSS~\cite{ilg2018occlusions} on the lakeside and sandbox scenes from the ETH3D~\cite{schoeps2017cvpr} two view stereo dataset.} 
	   	\label{fig:ETH}
	\end{figure*}
	
	\begin{figure*}
\centering			\includegraphics[width=\textwidth]{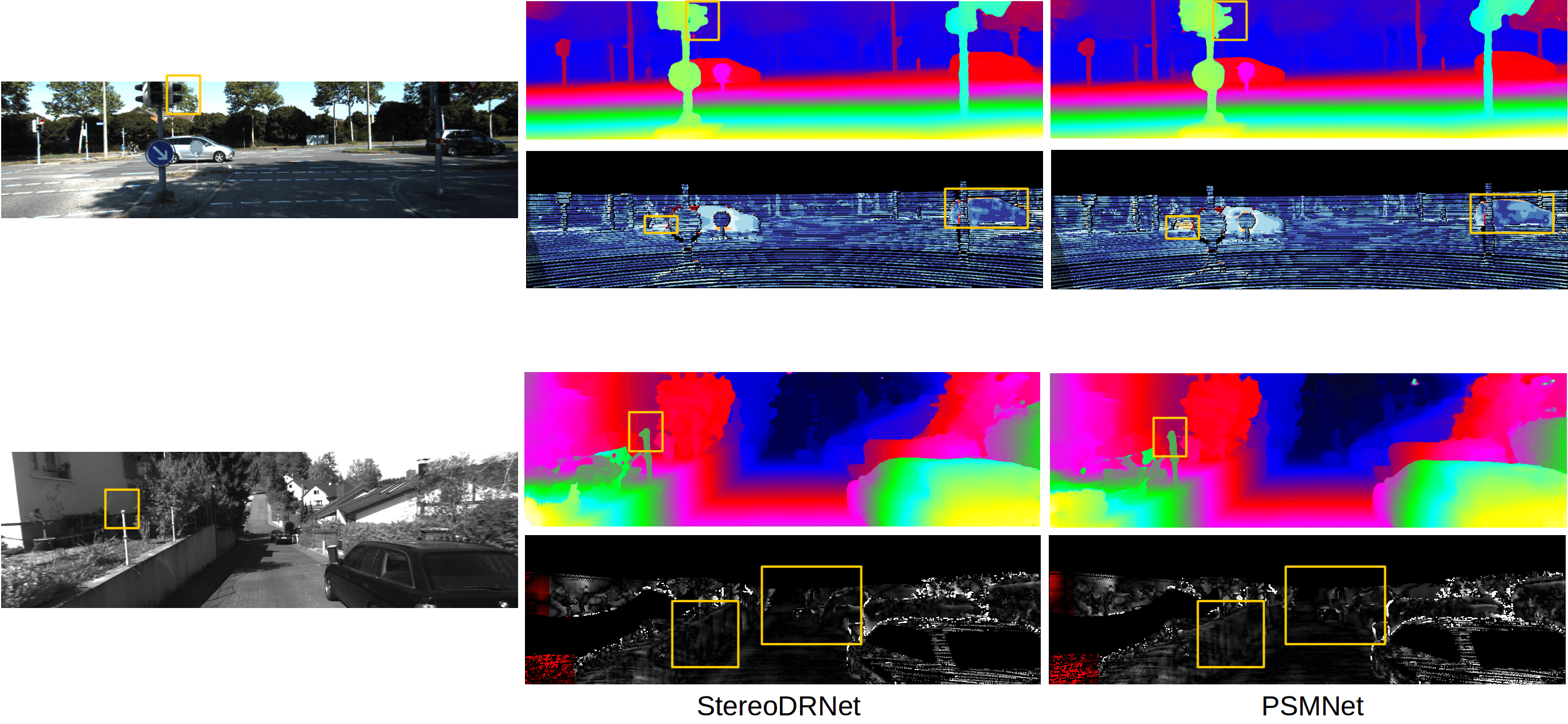}
	  	\caption{This figure shows the disparity estimation results of our StereoDRNet and PSMNet~\cite{chang2018pyramid} on the KITTI 2015 and the KITTI 2012 dataset. } 
	   	\label{fig:Kitti}
	\end{figure*}
    \clearpage
{\small
	\bibliographystyle{ieee}
	\bibliography{main}
}
\end{document}